\documentclass[letterpaper]{article}

\usepackage{microtype}
\usepackage{graphicx}
\usepackage{subcaption}
\usepackage{xcolor}
\usepackage{booktabs} 
\usepackage{authblk}

\usepackage{etoolbox}
\usepackage[a4paper, margin=1in, top=1.25in]{geometry}

\definecolor{darkblue}{rgb}{0,0,.75}
\definecolor{darkgreen}{rgb}{0,0.5,0}
\usepackage{hyperref}
\hypersetup{colorlinks=true,allcolors=darkblue}
\usepackage{macros}

\usepackage{enumitem}

\usepackage{amsmath}
\usepackage{amssymb}
\usepackage{mathtools}
\usepackage{amsthm}

\usepackage[capitalize,noabbrev]{cleveref}

\theoremstyle{plain}
\newtheorem{theorem}{Theorem}[section]
\newtheorem{proposition}[theorem]{Proposition}

\theoremstyle{definition}
\newtheorem{definition}[theorem]{Definition}
\newtheorem{assumption}[theorem]{Assumption}
\theoremstyle{remark}

\usepackage[textsize=tiny]{todonotes}

\usepackage{algorithm}
\usepackage{algpseudocode}
\usepackage{amssymb}
\usepackage{mathtools}
\usepackage{color}
\usepackage{comment}
\usepackage{algcompatible}
\newcommand*\samethanks[1][\value{footnote}]{\footnotemark[#1]}

\title{Differentially Private Heavy Hitter Detection using Federated Analytics}

\author[2]{Karan Chadha\footnote{This research was conducted while the author was an intern at Apple.}}
\author[1]{Junye Chen}
\author[1,2]{John Duchi}
\author[1]{Vitaly Feldman}
\author[1]{Hanieh Hashemi\thanks{Corresponding Authors: \texttt{h\_hashemi@apple.com}, \texttt{audra\_mcmillan@apple.com}}}
\author[1]{Omid Javidbakht}
\author[1]{Audra McMillan\samethanks}
\author[1]{Kunal Talwar}

\affil[1]{Apple Inc}
\affil[2]{Stanford University}
\date{}

\begin{document}

\maketitle

\begin{abstract} 
In this work, we study practical heuristics to improve the performance of prefix-tree based algorithms for differentially private heavy hitter detection. Our model assumes each user has multiple data points and the goal is to learn as many of the most frequent data points as possible across all users' data with aggregate and local differential privacy. \longversion{We 
propose an adaptive hyperparameter tuning algorithm that improves the performance of the algorithm while satisfying computational, communication and privacy constraints. We explore the impact of different data-selection schemes as well as the impact of introducing deny lists during multiple runs of the algorithm. We test these improvements using extensive experimentation on the Reddit dataset~\cite{caldas2018leaf} on the task of learning the most frequent words.}
\end{abstract}
\section{Introduction}
\label{sec:intro}
\looseness -1
Gaining insight into population trends allows data analysts to make data-driven decisions to improve user experience.
Heavy hitter detection, or learning popular data points generated by users, plays an important role in learning about user behavior.
A well-known example of this is learning ``out-of-vocabulary" words typed on keyboard, which can then be used to improve next word prediction models. 
This data is often sensitive and the privacy of users' data is paramount. When the data universe is small, one can obtain private solutions to this problem by directly using private histogram algorithms such as RAPPOR~\cite{erlingsson2014rappor}, and PI-RAPPOR~\cite{feldman2021lossless}, and reading off the heavy-hitters. However, when the data universe is large, as is the case with ``out-of-vocabulary" words, these solutions result in algorithms with either very high communication, or very high server side computation, or both.
Prefix-tree based iterative algorithms can lower communication and computation costs, while maintaining high utility by efficiently exploring the data universe for heavy hitters.
They also offer an additional advantage in the setting where users have multiple data points by refining the query in each iteration\longversion{ using the information learned thus far}, allowing each user to select amongst those data points which are more likely to be heavy hitters.

\looseness -1
In this work, we consider an iterative federated algorithm for heavy hitter detection in the aggregate model of differential privacy (DP) in the presence of computation or communication constraints. In this setting, each user has a private dataset on their device. In each round of the algorithm, the data analyst sends a query to the set of participating devices, and each participating device responds with a  \textit{response}, which is a random function of the private dataset of that user. These \textit{responses} are then summed using a secure aggregation protocol, and reported to the data analyst. The analyst can then choose a query for the next round adaptively, based on the aggregate results they have seen so far. The main DP guarantee is a user-level privacy guarantee on the outputs of the secure aggregator, accounting for the privacy cost of \emph{all} rounds of iteration. 
Our algorithm will additionally be DP in the local model of DP (with a larger privacy parameter)\footnote{A potential architecture for running iterative algorithms in this model of privacy is outlined in \cite{mcmillan2022private}.}.
\longversion{We do not assume that the set of participating devices is consistent between rounds. }

\looseness -1
In the central model of DP, there is a long line of work on adaptive algorithms for heavy hitter detection in data with a hierarchical structure such as learning popular $n$-grams~\cite{cormode2012differentially, Qardaji:2012, Song:2013, bagdasaryan2021towards, Kim2021DifferentiallyPN, mcmillan2022private}. These interactive algorithms all follow the same general structure. Each data point is represented as a sequence of data segments $d=a_1a_2\cdots a_r$ and the algorithm iteratively finds the popular values of the first segment $a_1$, then finds popular values of $a_1a_2$ where $a_1$ is restricted to only heavy hitters found in the previous iteration, and so on. 
This limits the domain of interest at each round, lowering communication and computation costs.
The method of finding the heavy hitters in each round of the algorithm varies in prior work, although is generally based on a DP frequency estimation subroutine. One should consider system constraints (communication, computation, number of participating devices, etc.) and the privacy model when choosing a frequency estimation subroutine. In this work, we will focus on using one-hot encoding with binary randomized response (inspired by RAPPOR~\cite{erlingsson2014rappor}) as our DP frequency estimation subroutine. Since we are primarily interested in algorithmic choices that affect the iterative algorithm, we believe our findings should be agnostic to the choice of frequency estimation subroutine used. 

We explore the effect on utility of different data selection schemes and algorithmic optimizations. 
We refer to our algorithm as \textit{Optimized Prefix Tree} (\textit{$\ouralgorithm$}). Our contributions are summarised below:

    \textbf{Adaptive Segmentation.} We propose an algorithm for adaptively choosing the segment length and the threshold for keeping popular prefixes. In contrast to prior works that treat the segment length as a hyperparameter, our algorithm chooses these parameters in response to user data from the previous iteration and attempts to maximize utility (measured as the fraction of the empirical probability distribution across all users captured by the returned heavy hitters), while satisfying any system constraints. We find that our method often results in the segment length varying across iterations, and outperforms the algorithm that uses a constant segment length. We also design a threshold selection algorithm that adaptively chooses the prefix list for the subsequent round. This allows us to control the false positive rate\longversion{ (the likelihood that a data point is falsely reported as a heavy hitter)}.
    
    \textbf{Analysis of the effect of on-device data selection mechanisms.} We explore the impact of interactivity in the setting where users have multiple data points. We observe empirically that when users have multiple data points, interactivity can improve utility, even in the absence of system constraints. In each round, users choose a single data point from their private data set to (privately) report to the server.
    The list of heavy hitters in the previous iteration provides a \emph{prefix list}, so users will only choose a data point with one of the allowed prefixes. If a user has several data points with allowed prefixes, then there are several selection rules they may use to choose which data point to report. Each user's private dataset defines an empirical distribution for that user. 
    We find that when users sample uniformly randomly from the support of their distribution (conditioned on the prefix list) then the algorithm is able to find more heavy hitters than when they sample from their empirical distribution (again conditioned on the prefix list). 

    \textbf{Analysis of the impact of inclusion of deny list.} Under the constraint of user-level differential privacy, each user is only able to communicate their most frequent data points, and less frequent data points are down weighted. We explore the use of a \emph{deny list} that asks users not to report data points that we already know are heavy hitters. In practice, a deny list may arise from an auxiliary data source, or from a prior run of the algorithm. Our analysis indicates even when the privacy budget is shared between multiple rounds of the algorithm, performing a second round equipped with a deny list improves performance.

The rest of the paper is organized as follows. In Section~\ref{sec:related} we discuss some of the prior works in privacy-preserving heavy hitters detection. Section~\ref{sec:privacy} explains the privacy primitives we used in this work. In Section~\ref{sec:alg} we elaborate the details of our prefix tree algorithm. Section~\ref{sec:post} explains the post-processing methods and the theoretical analysis behind it. Section~\ref{sec:expts} demonstrates the experimental results and in Section~\ref{sec:conclusion} we discuss the findings of our experiments. 

\longversion{\section{Related Works}\label{sec:related}
Heavy hitters discovery methods have applications in various different domains~\cite{elkordy2023federated}. This problem has been studied in both the local model~\cite{apple, wang2019locally, acharya2019hadamard} and shuffle model~\cite{ghazi2021power} of differential privacy. Furthermore, recently different multi-party-computing~\cite{boneh2021lightweight} methods and combination of multi-party-computing and DP techniques~\cite{bohler2021secure} have been proposed to find the top-k heavy hitters in different domains. In this work we focus on large domains and specifically iterative methods that allows us to satisfy system constraints. 

In~\cite{zhu2020federated}, the authors propose an iterative algorithm to discover heavy hitters in the central model of differential privacy. The general framework of forming a tree-based structure is the same to our Prefix Tree method except in their algorithm, $\triehhg$, samples a subset of \devices($\gamma \sqrt{\numusers}$) in each iteration and uses the data points of these \devices to compute the heavy hitters for the next iteration, without any additional noise and hence does not satisfy local differential privacy. They select the prefix list for the next iteration to be all the prefixes such that more than $\theta$ \devices send the character in that iteration. The parameters $\gamma$ and $\theta$ are chosen to achieve the required privacy guarantee. 

$\triehhp$~\cite{cormode2022sample} is an extension to $\triehhg$. The authors use the same sampling and threshold algorithm as $\triehhg$ to provide the $(\epsilon, \delta)-$aggregated differential privacy. However, they are able to support more general applications such as quantile and range queries. In addition to detecting heavy hitters, their method is able to report the frequency of heavy hitters without using additional privacy budget. To achieve their goal, they take advantage of Poisson sampling instead of fixed-size sampling to hide the exact number of samples. Consequently, releasing heavy hitters and their counts does not violate user privacy. 

Set Union is a critical operation in many data related applications. Differentially Private Set Union (DPSU) methods are~\cite{gopi2020differentially} popular for extracting n-grams, which is a common application in NLP algorithms. These methods attempt to find the largest subset of the union while satisfying DP. Authors in~\cite{wilson2020differentially}, samples a specific number of items per user and generates a histogram. Finally the items whose counts are above a certain threshold will be reported. In~\cite{carvalho2022incorporating}, utility is boosted by privately reporting the frequencies to the server and eliminating the sampling step. They further take advantage of knowledge transfer from public datasets to achieve more accurate frequency estimation. Using public data as prior knowledge for private computation is investigated in various other works~\cite{liu2021leveraging, bassily2020learning}. In this paper, we use this knowledge transfer to explore the effect of using a  deny list on the utility of the algorithm. Authors in~\cite{Kim2021DifferentiallyPN} combined DPSU and tree based method to improve the utility of n-gram extraction model. The empirical results of their work imply that selecting more than one data point per device improved performance in the central DP setting. 
While we focus on data selection mechanisms that select a single data point per user per round, these mechanisms naturally extend to mechanisms that select more than one data point. In order to elaborate the impact of weighted vs. unweighted sampling, we focus on selecting a single data point per device. We leave an exploration of the optimal number of data points per device per iteration in the aggregate DP setting to future work. 
}

\longversion{\section{Differential Privacy}\label{sec:privacy}

In this work, we will consider an algorithm that satisfies \devicenospace-level differential privacy (DP) in the aggregate model \emph{and} the local model of differential privacy. 
We focus on \devicenospace-level DP, which protects against a user changing all of the data points associated to them. Our primary privacy guarantee is the aggregate privacy guarantee, which will be specified ahead of time. \longversion{Local differentially private guarantees are achieved locally on a user's device through the use of a local randomizer.}
The local privacy guarantee will be set to be the largest epsilon such that the final algorithm satisfies the required aggregate privacy guarantee (i.e. we will not put constraints on the local privacy guarantee).

Local differentially private guarantees are achieved locally on a user's device through the use of a local randomizer.

\begin{definition}[Local Randomizer \cite{DR14,kasiviswanathan2011can}]\label{LDP}
Let $\A: \worddom \to \cY$ be a randomized algorithm mapping a data entry in $\worddom$ to an output space $\cY$.  
The algorithm $\A$ is an $\epsilon$-DP local randomizer if for all pairs of data entries $d,d'\in\worddom$, and all events $E\subset\cY$, we have $$
- \eps \leq \ln\left(\frac{\Pr[\A(d) \in E ]}{\Pr[\A(d') \in E ]}  \right)\leq \eps.
$$
\end{definition}
The privacy parameter $\eps$ captures the \emph{privacy loss} consumed by the output of the algorithm. Differential privacy for an appropriate $\eps$ ensures that it is impossible to confidently determine what the individual contribution was, given the output of the mechanism. 

In general, differential privacy is defined for algorithms with input databases with more than one record. In the local model of differential privacy, algorithms may only access the data through a local randomizer so that no raw data leaves the device. For a single round protocol, local differential privacy is defined as follows:

\begin{definition}[Local Differential Privacy \cite{kasiviswanathan2011can}]\label{localDP}
Let $\A: \worddom^n \to \cZ$ be a randomized algorithm mapping a dataset with $n$ records to some arbitrary range $\cZ$.  The algorithm $\A$ is $\epsilon$-local differentially private if it can be written as $\A(d^{(1)}, \cdots, d^{(n)}) = \phi\left(\A_1(d^{(1)}), \cdots, \A_n(d^{(n)}) \right)$ where the $\A_i: \worddom \to \cY$ are $\epsilon$-local randomizers for each $i \in [n]$ and $\phi: \cY^{n} \to \cZ$ is some post-processing function of the privatized records $\A_1(d^{(1)}),\cdots, \A_n(d^{(n)})$.  Note that the post-processing function does not have access to the raw data records.
\end{definition}

We say a multi-round algorithm $\A$ is $\eps$-DP in the local model if it is the composition of single round algorithms which are DP in the local model, and the total privacy loss of $\A$ is $\eps$-DP. More generally, we can say that an interactive algorithm is locally differentially private if the transcript of all communication between the data subjects and the curator is differentially private~\cite{joseph2019role}. Since aggregate differential privacy is our primary privacy guarantee, when we refer to local privacy guarantees, they will be for a single round of communication.

\longversion{In aggregate DP, we assume the existence of an aggregation protocol that sums the local reports before they are released to the analyst. The aggregation protocol guarantees that the analyst does not receive anything about the locally DP reports \emph{except their sum}\footnote{The aggregate model of DP is a derivative of the more general and common shuffle model of differential privacy introduced in \cite{ErlingssonFMRTT19, CheuSUZZ19}.}.} 

\begin{definition}\label{defnaggregateDP}
A single round algorithm $\A$ is $(\eps, \delta)$-DP in the aggregate model if the output of the aggregation protocol on two datasets that differ on the data of a single individual are close. Formally, an algorithm $\A:\cD^n\to\cZ$ is $(\eps, \delta)$-DP in the aggregate model if the following conditions both hold:
\begin{itemize}
\item it can be written as $\A(d^{(1)}, \cdots, d^{(n)})=\phi(\aggregator(f(d^{(1)}), \cdots, f(d^{(n)}))$ where $f:\cD\to\cZ$ is a randomized function that transforms that data, $\aggregator$ is an aggregation protocol, and $\phi:\cY^n\to\cZ$ is some post-processing of the aggregated report
\item for any pair of datasets $D$ and $D'$ that differ on the data of a single individual, and any event $E$ in the output space,
\[
\Pr(\A(D)\in E)\le e^{\eps} \Pr(\A(D')\in E)+\delta.
\]
\end{itemize}
Note that the post-processing function takes the aggregation as its input and does not have access to the individual reports.
\end{definition}

\longversion{When each user uses a local randomizer with a local DP guarantee to send their data to the aggregation protocol, the privacy guarantee in the aggregation model can be bounded by a quantity that is a function of both $\epslocal$, the privacy guarantee in the local model, and $n$, the number of users that participate in the aggregation protocol~\cite{ErlingssonFMRTT19, CheuSUZZ19}.} In our experimental results, we will bound the aggregate DP epsilon by the numerical bound on privacy amplification by shuffling due to \cite{feldman2023stronger}. 
Given an expected number of users, the number of iterations, and the desired aggregate privacy guarantee, we solve for the largest local epsilon that will achieve the given aggregate privacy guarantee. Since the aggregate privacy guarantee is our primary privacy guarantee, we do not put an upper bound on our local epsilon.  

As with the local model, we will say a multi-round algorithm $\A$ is $\eps$-DP in the aggregate model if it is the composition of single round algorithms which are DP in the aggregate model, and the total privacy loss of $\A$ is $\eps$-DP.
In order to analyse the privacy loss over multiple iterations, we will use a combination of the advanced composition theorem~\cite{DRV, Kairouz:2017} and composition bound in terms of R\'enyi differential privacy~\cite{DLDP, mironov2017renyi, canonne2020discrete}.
\longversion{Table~\ref{tab:eps} in Appendix~\ref{appendix:privacy} demonstrates how the local epsilon increases with the number of iterations, and the desired aggregate privacy guarantee.}}

\section{Algorithm}\label{sec:alg}
In this section, we describe our proposed algorithm $\privhh$. \longversion{This algorithm privately identifies the heavy hitters using $\numiters$ iterations where each \device sends a locally differentially private report in each round and the local reports are aggregated before reaching the server. We will first discuss an outline of the high-level algorithm.}
In Section~\ref{sec:post}, we will discuss our proposals for adaptively setting the various parameters and subroutines present in the high-level algorithm. Our focus in the experimental section to follow will be to explore these choices, and provide some guidelines on how they should be chosen. Note that while our algorithm is run over multiple iterations, it is well-suited to the federated setting since it does not require every user to be present at every iteration.

We represent the system constraints as a constraint of the size of the data domain for any single iteration, denoted by $\complimit$. This bound may be a result of communication constraints, as is the case for the local randomizer we will use (one-hot encoding with binary randomised response), or computational constraints for the server-side algorithm, as in PI-Rappor~\cite{feldman2021lossless} or Proj-Rappor~\cite{feldman2022private}.

\textbf{Notation.}
\longversion{Let $\numusers$ be the total number of users. }\longversion{Let each user $i\in[N]$ have $\ndp_i$ data points 
denoted by $\datap_{i,1},\dots,\datap_{i,n_i}$ from domain $\worddom \subset  \worduniv = \alphadom^r$, where $\worduniv$ denotes the universe of allowable data points, and $\alphadom$ denotes an alphabet from which all data points are built.} For each user $i\in[N]$, let $\empuserdist_i$ denote the empirical distribution of user $i$'s data and $\empuserdist~\coloneqq~\frac{1}{\numusers}\sum_{i \in [\numusers]}\empuserdist_i$ denote the global empirical distribution
\longversion{\footnote{In the multiple data points per \device setting, there are several other natural ways of defining the global empirical distribution. For example, one may define the frequency of a data point $d$ to be the number of users who have the word $d$ in their support. We briefly explore this metric in Appendix~\ref{appen:expts-details}.}}. Let each data point $\datap\in \alphadom^r$ be of a fixed length $\totlen$.

\subsection{Private Heavy Hitters Algorithm}

\longversion{$\privhh$ proceeds in iterations with the goal of efficiently exploring the large data domain to detect heavy hitters, by exploiting the hierarchical structure by sequentially learning the most popular prefixes, and only expanding on these popular prefixes in the next iteration.} \longversion{We give pseudo-code for our proposed algorithm $\privhh$ in \Cref{alg:privhh}.}
At every iteration $t$, the server sends the devices a list of live prefixes $\prefixlist_t$ of length $\preflent$, a deny list $\denylist$, and a segment length $\seglent$. The devices then use a data selection mechanism to choose a data point that is not in the deny list $\denylist$ and whose $\preflent$ length prefix belongs in $\prefixlist_t$, they then (privately) report back the length $\preflent+\seglen_t$ $(\leq r)$ prefix of the chosen data point. The server uses these local reports to define  $\prefixlist_{t+1}$ (consisting of prefixes of length $\preflent+\seglen_t$) and the segment length $\seglen_{t+1}$ for the next round.  
We will use $\numiters$ to denote the number of iterations and $\epslocal$ to be the local DP parameter for a single iteration. At the end of $\numiters$ rounds, our algorithm outputs a set of heavy hitters that includes the prefixes found in the last iteration ($\prefixlist_{T+1}$) and the contents of $\denylist$. We use $(\eps_{\rm agg}, \delta)$-DP to refer to the privacy parameters of our algorithm in the aggregate model. 

\longversion{We may also remove some prefixes during earlier iterations which we add into the final set of heavy hitters. These data points are stored in the ``discovered list", $\discoveredlist$. We'll discuss this further in a subsequent section, but this turns out to be a useful improvement when the natural encodings of data points can vary in length. For example, when using Huffman encoding.}

\begin{algorithm}
\caption{\label{alg:privhh} Prefix tree based heavy hitter algorithm $(\privhh)$}
\begin{algorithmic}[1]
\STATE \textbf{Input}: $\numiters$: number of iterations, $\epslocal$: Local privacy parameter, $\datasamp$: Data selection mechanism, $\complimit$: Bound on the dimension, $\falseratio$: false positive ratio, $\eta$: extra parameters to pass to $\serverper$, $\denylist$: deny list.
\STATE \textbf{Output}: $\prefixlist_{T+1}$: Set of Heavy Hitters
\STATE $\seglen_1 \gets \lfloor \log(\complimit) \rfloor$, $\prefixlist_1=\emptyset, \discoveredlist=\emptyset$ \textit{{\color{darkgreen} //  Initialize segment length and prefix list }}
\FOR{$t \in [T]$}
\STATE $\usercompdataset_t \gets \emptyset$ \textit{{\color{darkgreen} //  $V_t$ will be the set of all the device responses that is sent to the aggregation protocol }} 
\FOR{$i \in [\numusers]$}
\STATE $\usercompdata_i \gets \devicehh_i(\epsilon_l, \seglen_t, \prefixlist_{t}, \denylist, \datasamp)$
\STATE $\usercompdataset_t \gets \usercompdataset_t \cup \usercompdata_i$
\ENDFOR 
\STATE $V_t \gets \texttt{AggregationProtocol}(V_t)$ \textit{{\color{darkgreen} //  Device responses are aggregated }}
\STATE $\queryset_t \gets \prefixlist_{t} \times \alphadom^{\seglen_t}$ \textit{{\color{darkgreen} // Data domain for iteration $t$ }}
\STATE $\prefixlist_{t+1}, \seglen_{t+1}, \discoveredlist \gets \serverper(\usercompdataset_t, \queryset_t, \discoveredlist, \epsilon_l, \falseratio, \eta)$ \textit{{\color{darkgreen} //  Server returns prefix list and segment length for next round }}
\STATE Send $\mathcal \prefixlist_{t+1}$, and $\seglen_{t+1}$ to all the devices
\ENDFOR
\STATE \textbf{return} $\prefixlist_{T+1} \cup \denylist \cup \discoveredlist$
\end{algorithmic}
\end{algorithm}

\subsection{Device Algorithm} 

The device first uses the data selection mechanism \datasamp~to choose a data point from their on-device dataset that is not in the deny list, and whose $\preflent$-length prefix is in $\prefixlist$. Then we pass the $\preflent + \seglen_t$-length prefix of the chosen data point to a $\epslocal$-local DP algorithm\longversion{ $\compressor$} and send the privatized output to the aggregation protocol. \Cref{alg:client} gives more details for the device-side algorithm. 

\paragraph{The Local Randomizer} In our experiments we use one-hot encoding with asymmetric binary randomized response (denoted by \OHEBRRnospace) as the local randomizer. For details of this randomizer, see Appendix A of \cite{mcmillan2022private}.\longversion{ In practice one could use PI-RAPPOR~\cite{feldman2021lossless} or Proj-RAPPOR ~\cite{feldman2022private} for better communication-computation trade-offs (See \Cref{appen:pi-rappor}). Since the utility guarantees of these mechanisms are very similar to \OHEBRRnospace, we expect our findings on \OHEBRR to be directly applicable when using PI-Rappor or Proj-Rappor.}
\longversion{

Under the constraint of local differential privacy, a participating \device must still send a local report to the aggregation protocol, even if the \device data contains no data points with a prefix in the prefix list. In our experiments if a \device  has no data point to communicate, then it encodes this as the all-zeros vector and uses asymmetric randomized response on the coordinates of the all-zeros vector. This has the same effect as adding a special element $\perp$ to the data domain and having devices report the $\perp$ element if they have no data points to report.}

\paragraph{Data Selection} We consider two data selection mechanisms. 
In \emph{weighted selection}, each device $i$ selects a data point by sampling from its empirical distribution $\empuserdist_i$ conditioned on the datapoint having a prefix in $\prefixlist_{t}$ and not being in $\denylist$.
\longversion{Formally, we define the pdf of $\empuserdist_i$ as follows: let $f_i(\datap_j)$ represent the frequency of $d_j$ in the private data set on $device_i$ so $\sum_{j=0}^{u_i} f_i(\datap_j) = 1$ where $u_i$ is the number of unique data points on $device_i$.
In \emph{unweighted/uniform selection}, each device $i$ selects a data point by sampling uniformly from those points in the support of $\empuserdist_i$ which have a prefix in $\prefixlist_{t}$ and are not in $\denylist$. Note that the data selection mechanism does not impact the privacy guarantees.
$\datap_{i,1}$}

\begin{algorithm}
\caption{\label{alg:client} Device side algorithm $(\devicehh)$}
\begin{algorithmic}[1]
\STATE \textbf{Input:} $\epslocal$: Local privacy parameter, $\preflen$: prefix length, $\seglen_t$: segment length, $\prefixlist$: allowed prefix list, $\denylist$: deny list, $\datasamp$: Function to choose a datapoint from the data 
\STATE \textbf{Param:} $\dataset$: \device dataset
\STATE \textbf{Output:} $v$: Privatized output
\STATE $\dataset = \{\datap \in \dataset \mid \datap[0:\preflen] \in \prefixlist \wedge  \datap \notin \denylist\}$
\IF{$\dataset == \phi$} 
    \STATE $\datap \gets \perp$ \textit{{\color{darkgreen} // We reserve a special data element for users that have no eligible data points to report. }}
\ELSE
\STATE $\datap \gets \datasamp(\dataset)$
\ENDIF

\STATE $\usercompdata \gets \compressor(\datap[0:{\preflen + \seglen_t}]; \epsilon_l)$ 
\STATE \textbf{return} $\usercompdata$
\end{algorithmic}
\end{algorithm}

\subsection{Server Algorithm}
The server receives the aggregated privatized responses\longversion{ from the previous iteration, the prefix list and segment lengths from the previous iteration, and the local DP parameter $\epslocal$}. The general outline of the server-side algorithm is given in \Cref{alg:server-per}.

The first step of this process is to compute an estimated frequency for the data domain of the last iteration $\prefixlist_{t}\times A^{\seglent}$.
Given the aggregated privatized results \longversion{and the local epsilon, for every element, $d\in \prefixlist_{t} \times A^{\seglent}$, of the data domain,} the server can compute an estimate $\freqest(d)$ of the number of devices who sent the data point $d$ in the last iteration. The prefix list selection algorithm aims to keep as many of the elements\longversion{ $d\in \prefixlist_{t} \times A^{\seglent}$} such that $\freqest(d)>0$ as possible, while minimizing the number of ``false positives" in the prefix list (data elements which do not match the selected data point for \emph{any} device).
When using \OHEBRRnospace, each estimate $\freqest(d)$ is unbiased and the noise induced by the privatization scheme is approximately Gaussian with standard deviation $\sigma$, where $\sigma$ is a function of $\epslocal$ and the number of participating devices. Due to the noise, if we were to define the $\prefixlist_{t+1}$ to be all elements such that $\freqest(d)>0$, the false positive rate would be too high. 
Instead, we use a threshold multiplier $\tau$ such that the prefix list $\prefixlist_{t+1}$ contains all the elements such that $\freqest(d)\ge \tau\sigma$. This threshold should be chosen to be as small as possible while ensuring that the fraction of reported elements that are false positives does exceed a specified threshold denoted $\falseratio$.

\longversion{In the setting where there are natural encodings of the data for which different data points have different lengths (e.g. \longversion{when an encoding scheme such as }Huffman encoding\longversion{ is used}) we note a improvement that can be made when choosing the prefix list $\prefixlist_{t+1}$. In these cases, some data points may be complete before the end of the algorithm.}
\longversion{To avoid unnecessary communication and utilize the system capacity, an end character symbol can be used at the end of each encoding so that the server can detect when a data point is ``complete". After aggregating the data on the server if there are prefixes that reach the end character, they are \longversion{added to the list of already discovered prefixes and }removed from the prefix list sent to the devices for the next iteration. They are then added to the set of heavy hitters in the final output.}

In most of the prior works, the segment length and threshold $\tau$ are treated as hyperparameters that need to be tuned. Tuning hyperparameters is notoriously hard in the federated setting. In Section~\ref{sec:post} we will discuss our adaptive algorithms for choosing these parameters.
Our proposed algorithms choose these parameters in response to user data, without using additional privacy budget.

\begin{algorithm}
\caption{\label{alg:server-per} Server side algorithm per round $(\serverper)$}
\begin{algorithmic}[1]
\STATE \textbf{Input}: $\usercompdataset_t$: Aggregated sum of devices responses, $\queryset_t$: Data domain of iteration $t$, $\epslocal$: Local privacy parameter, $\falseratio$: False positive ratio, $\dpmult_0$: Initialization of threshold and $\eta$: Extra parameters for $\prunehh$
\STATE \textbf{Output}: $\prefixlist$: Heavy hitters list 
\STATE $\freqest(\cdot) \gets \decompressor(\usercompdataset_t,\queryset_t; \epslocal)$ \textit{{\color{darkgreen} // Takes the aggregated privatized responses and computes an estimate of the frequency of every data element. }}
\STATE $\sigma \gets \sqrt{\var (\decompressor)}$ \textit{{\color{darkgreen} // Computes an upper bound on the standard deviation of the frequency estimate for $d$ }}
\STATE $\prefixlist_{t+1} \gets \prunehh(\queryset,\freqest(\queryset), \dpmult_0, \falseratio, \sigma, \eta)$ 
\STATE $\prefixlist_{t+1}, \discoveredlist \gets \texttt{RemoveFinished}(\prefixlist_{t+1},\discoveredlist)$ \textit{{\color{darkgreen} // Removes any of the discovered prefixes which are "complete" data points and adds them to the discovered list. }}
\STATE $\seglen_{t+1}=\max\{\ell\;|\; |\prefixlist_{t+1}|\times 2^{\ell}\le \complimit\}$ \textit{{\color{darkgreen} // Adaptively chooses the maximum segment length }}
\STATE \textbf{return} $\prefixlist_{t+1}, \seglen_{t+1}, \discoveredlist$
\end{algorithmic}
\end{algorithm}

\section{Adaptive Thresholding and Segmentation}\label{sec:post}

In this section, we will describe our adaptive segmentation and thresholding algorithms. The algorithms aim to keep as many heavy hitters as possible, while maintaining a specific false positive rate, and satisfying the data domain size constraint for the next iteration.

\subsection{Adaptive Thresholding}

Let us first discuss our proposal for how to choose the threshold $\tau$.
Given a threshold $\tau$ and standard deviation $\sigma$, let $E$ denotes the probability that a data point with true count zero (i.e. no device contributes this data point) has an estimated count above $\tau\sigma$. That is, the probability of a false positive. As discussed earlier, for any $d$, the estimate $\tilde{f}(d)$ is approximately Gaussian with standard deviation $\sigma$ so we can compute $E$ based on the Gaussian approximation, i.e. $E = 1-\Phi(\tau\sigma)$, where $\Phi()$ denotes the gaussian CDF (with mean 0 and standard deviation $\sigma$)\footnote{if $d$ has true count 0, then the distribution of $\tilde{f}(d)$ is actually a shifted, scaled Binomial distribution. If $n$ is small enough that the Gaussian approximation is not accurate, then one can use high probability bounds on the Binomial.}. 

In the while loop (line 4 to 7), we first compute the expected number of false positive bins by $|\mathcal{D}|\times E$ where $|\mathcal{D}|$ represents the aggregated data domain size. We then make sure that the ratio of the expected number of false positives to the number of data points such that $\tilde{f}(d)\ge\tau\sigma$ (represented by $|\prefixlist'|$) does not go above a specified threshold. If the ratio of the expected number of false positives that threshold $\tau$ to the number of data elements with estimated frequency above $\tau\sigma$ exceeds the specified false positive ratio, we change the confidence level to the point that we make sure the algorithm satisfies this parameter.

\begin{algorithm}[H]
\caption{\label{alg:prunehh}Pruning algorithm for confident heavy hitter detection ($\prunehh$)}
\begin{algorithmic}[1]
\STATE \textbf{Input}: $\queryset,\freqest[\queryset]$: query set and estimated frequencies after aggregation, $\dpmult_0$: Initialization of threshold, $\falseratio$: ratio of expected false positives to the total number of bins, $\sigma$: aggregated noise standard deviation, $\eta$: step size

\STATE $E \gets 1-\Phi(\dpmult_0\sigma)$ 
\STATE  {$\prefixlist'= \{q \in \queryset ~\mid~ \freqest[q] >\dpmult_0 \times \sigma\} $}
{\WHILE {${\falseratio} < \frac{E \times |\mathcal{D}|}{|\prefixlist'|} $} 
\STATE $E \gets \eta*E$
\STATE $\dpmult \gets z_{(1-E)}$ 
\STATE  {$\prefixlist'= \{q \in \queryset ~\mid~ \freqest[q] >\dpmult \times \sigma\} $}
\ENDWHILE}

\STATE {\textbf{return} $\prefixlist$}

\end{algorithmic}
\end{algorithm}

\subsection{Adaptive Segmentation}\label{sec:analysis}\label{sec:adaptive}

Given the prefix list, the segment length is adaptively chosen to be as large as possible while maintaining the dimension constraint, $\seglen_{t+1}~=~\arg\max_{\ell}\{\ell\; |\;|\prefixlist_{t+1}| \cdot 2^{\ell}\le \complimit \}$. 
In the single data point per device setting, intuitively, there are two opposing factors in the performance of the private heavy hitters algorithm --- the privacy budget per iteration (which decreases with increase in $\numiters$) and the size of the total search space (which is smaller for algorithms with smaller segmentation lengths and hence more iterations). In this section, we outline an argument illustrating that the effect of decreasing in privacy budget per iteration dominates and it is better to minimize the number of iterations. We also illustrate this via experiments. Thus, we choose each segment length to be as large as possible while retaining as many popular prefixes (with suitable confidence) as possible and maintaining the dimension constraint. 

\paragraph{Intuitive theoretical analysis of $\privhh$ for single datapoint}
In this part, we try to obtain guidelines for how to set the segment length in the a single data point per device setting. We provide analysis for running $\privhh$ for one round ($T = 1$) searching over the whole high dimensional universe $ \alphadom^\totlen$. While this may be impractical to implement, it provides us with setting of hyperparameters in the case of no computation and communication constraints. We analyze the algorithm assuming good performance of the local randomizer $\compressor$, frequency estimator $\decompressor$ pair. We formalize this assumption as follows: 

\begin{assumption}\label{ass:local-rand-perf} For any $\beta\in [0,1]$, for any element $x$ in the domain,
with probability at least $1 - \beta$, we have, \[|F(x) - \freqest(x)| \leq C_1\sqrt{\frac{ne^\epslocal}{(e^\epslocal - 1)^2}\log\left(\frac{1}{\beta}\right)}=C_2\frac{\sqrt{\log(1/\delta)\log(1/\beta)}}{\epsilon_{agg}},\] where $C_1$ and $C_2$ are absolute constants, $F$ is the global empirical distribution and $\tilde{f}$ is the estimated empirical distribution. 
\end{assumption}

We note that this assumption is satisfied by \OHEBRRnospace, as well as other common frequency estimation algorithms such as PI-Rappor. For the purpose of this analysis, we will use a different metric to the metric that will be the main focus in our experimental results. We will measure the performance of $\privhh$ using a metric we define in \Cref{def:lam-acc}, which is standard in the differentially private heavy hitters literature \cite{bassily2017practical}. 

\begin{definition}[$\lambda$-accurate]\label{def:lam-acc}
A set of heavy hitters $\prefixlist_\numiters$ is said to be $(\lambda, A)$-accurate if it satisfies the following:

\begin{itemize}
\setlength\itemsep{1mm}
    \item For all $d \in \worduniv$, if $F(d) \geq A + \lambda$, then $d\in \prefixlist_\numiters$.
    \item For all $d \in \worduniv$, if $F(d) < A - \lambda$, then $d \notin \prefixlist_\numiters$.
\end{itemize}
\end{definition}

We now prove the utility of $\privhh$ when we have one iteration with $\seglen = \totlen$ in \Cref{prop:one-round}.
We first state the result and discuss its implications, deferring the proof to Appendix~\ref{appendix:proof}. 

\begin{proposition}\label{prop:one-round} 
    Let the local randomizer ($\compressor$) and frequency estimator ($\decompressor$) pair satisfy \Cref{ass:local-rand-perf} and let $\prefixlist_1$ be the output of $\privhh$ when run for $T = 1$ round with $\seglen = \totlen$ and the query set $\queryset = \alphadom^\totlen$ and aggregate DP parameters $({\epsilon_{agg}}, \delta)$. Then, with probability at least $(1-\beta)$, $\prefixlist_1$ is $(\lambda, \tau\sigma)$-accurate with $\lambda = O\prn{\frac{1}{\epsilon_{agg}}\sqrt{\log(1/\delta)\log\prn{\frac{|\alphadom|^\totlen}{\beta}}}}$, where $\tau\sigma$ is the final threshold.
\end{proposition}

While this only give us upper bound on the performance of the single iteration algorithm, it does help us gain some intuition into how to set the segmentation length. Suppose we were to run the algorithm for $T$ iterations with even segmentation (i.e. $\seglen_t=r/T$ for all $t$). Then, at each iteration, we would need the aggregate privacy guarantee for that round to be $\approx\epsilon_{agg}/\sqrt{T}$. If the data domain size per iteration was exactly $|A|^{r/T}$ then the impact of $T$ on the error would approximately cancel (ignoring logarithmic terms that arise from ensuring true heavy hitters survive at each iteration). However, the data domain at each round is \emph{greater} than $|A|^{r/T}$ pushing us towards using a single round.

This intuition does not generalize to the multiple data points per device setting as it is possible for iterations allows us to more intelligently select the data points that users contribute. This is aligned with our empirical results in Appendix~\ref{appendix:multioptprefixtree} which indicates sending one character at a time improves the utility. However, we conjecture that reducing the number of iterations per round, and including multiple rounds with deny-lists will improve utility.

\section{Experiments}\label{sec:expts}

\begin{figure*}
\centering
    \begin{subfigure}[b]{0.45\textwidth}
        \includegraphics[width = \textwidth]{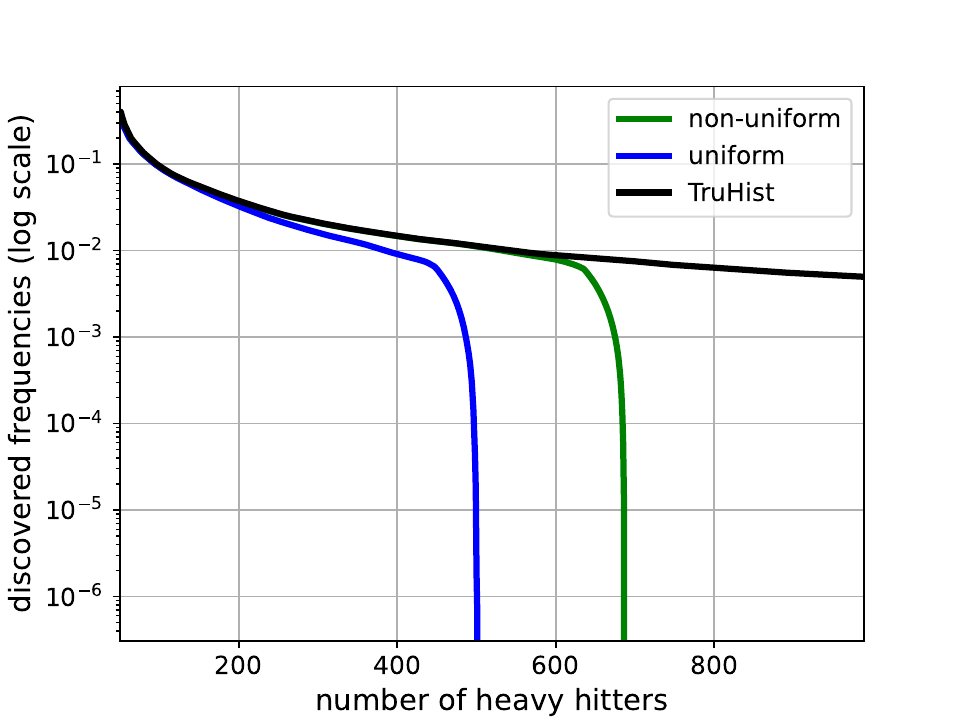}
        \caption{Discovered frequencies for different segmentation schemes in $\ouralgorithm$}
        \label{fig:uniformvsnon}    
    \end{subfigure}
    \hfill
    \begin{subfigure}[b]{0.45\textwidth}
             \centering
             \includegraphics[width = \textwidth]{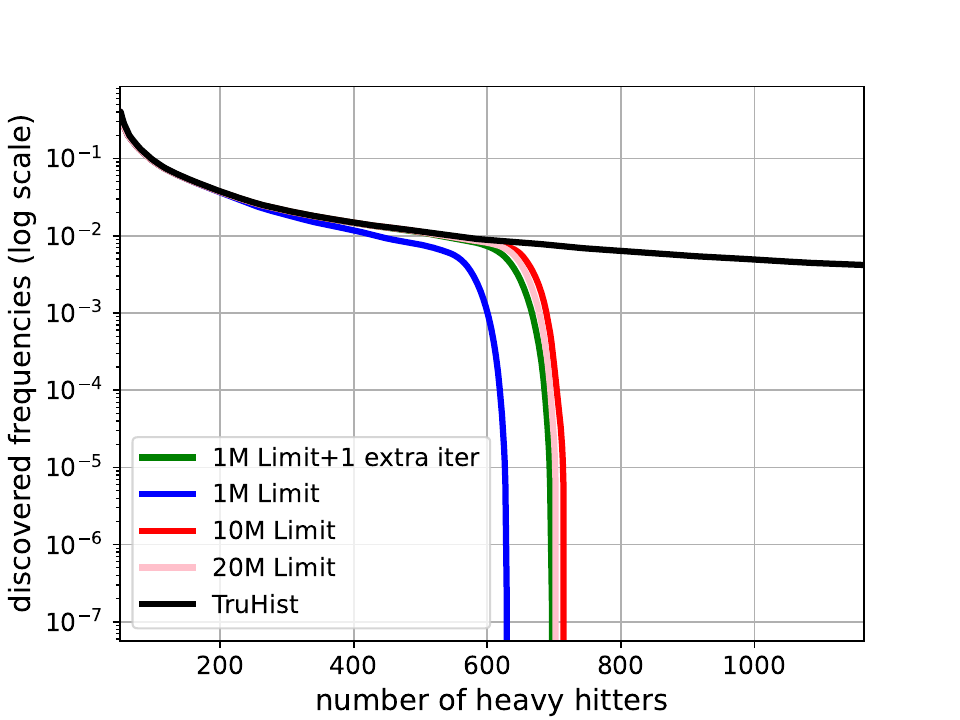}
         \caption{Discovered frequencies for different dimension limits in $\ouralgorithm$ }
        \label{fig:PayloadLimitFreq}
         \end{subfigure}
    \caption{Experimental results in the single data point per \device setting with $\epsagg=1$, $\delta=10^{-6}$, and $\numiters = 4$}
\end{figure*}
\textbf{Evaluation Dataset:} For our evaluations, we use the Reddit public dataset which contains the data of $1.6$ million users' comments posted on social media in December 2017~\cite{caldas2018leaf}. On the Reddit dataset each device has an average of $1092$ words and an average of $379$ unique words.
For investigations on the single data point per \device setting, each \device $i$ samples a single data point from $\empuserdist_i$. This data point remains fixed during the multiple iterations of the algorithm. 
Unless stated otherwise, we use a Huffman encoding to translate the words into binary strings. One token is reserved for unknown characters and we have an end character encoding at the end of each bit stream. We set the system constraint $\complimit=10^7$ and $\totlen = 60$ in all experiments.

\paragraph{\textbf{Evaluation metric:}}Let $H = (x_1,x_2,\dots,x_{|H|})$ denote the set of heavy hitters output by an algorithm ordered by the empirical global frequency distribution $\empuserdist$. For our evaluations in this section, we report the frequencies (according to $\empuserdist$) of heavy hitters output by the algorithm. To aid with visualization in our plots, for a window size $W=50$ and a given heavy hitter set $H$, we plot for each $i$, the sum of the probabilities (according to $\empuserdist$) of the heavy hitters in the sliding window 
(more details on plots are in Appendix~\ref{appendix:tuning}).

\subsection{Adaptive Segmentation in Single Data Point Setting}
\label{sec:hyper}

In this section, we focus on the simpler single data point per \device setting. \longversion{For these evaluations we used $\epsagg=1$ and $\delta = 10^{-6}$.}  
 
 \paragraph{Adaptive Segmentation}
 
We demonstrate the benefit of adaptively choosing the segment length, as opposed to using fixed-length uniform segment lengths.  \longversion{For our experimental evaluations in this part we set our data domain limit to $\complimit = 10^7$ which means that the dimension of each payload of size $2^{\seglen_{t}} \times |\prefixlist_t| $ should not exceed this limit. For both of the configurations in this part, we limit $\numiters=4$.} \longversion{In one configuration we used the fixed segment length of 15 for all the iterations. On the other hand, with $\ouralgorithm$ in each iteration we select the largest possible segmentation length for the next iteration based on the dimension limitation and the size of the {prefix list} for the next iteration.} 
Fig.~\ref{fig:uniformvsnon} compares the discovered normalized frequencies for the algorithm which uses uniform segment lengths ($\seglen_t=15$ for all $t$) to the algorithm which selects the segments adaptively, which leads to segment lengths [23, 14, 11, 12]. Our adaptive scheme is able to discover $40\%$ more heavy hitters (more plots in Fig.~\ref{fig:segmentation}). We use adaptive thresholding in both algorithms, with $\falseratio = 0.5$. However, the empirical $\falseratio$ is even lower with the values of $0.35$ and $0.41$ for adaptive and uniform segment algorithms, respectively.

 \paragraph{Dimension Limitation}
 
 In production-scale systems the computation cost of decompressing and aggregating the responses can be a significant bottleneck when the data comes from a very large domain. Increasing system limits can be expensive and resource intensive. Thus, we explore the effect of different dimension constraints on the performance of $\ouralgorithm$. In order to do so, we used different constraints of $P=$20, 10, and 1 million. This limitation in iteration $t$ specifies an upper bound on $2^{\seglen_{t+1}} \times |\prefixlist_t|$. First for all the configurations we set the number of iterations to 4. As illustrated in Figure~\ref{fig:PayloadLimitFreq} changing the limit to 1 million degrades the utility of the algorithm significantly. This is because the small number of iterations and small system limit means that in the final round, the prefix list has to be smaller than desired in order to set the segment length large enough to finish the algorithm in the desired number of iterations. One way to compensate this degradation is to allow the algorithm to run in 5 iterations instead of 4. To account for the total privacy budget in all the iterations, by having one more iteration, the per iteration local epsilon $\epslocal$ decreases (check Table~\ref{tab:eps}). However, with 5 iterations, the algorithm is able to detect $15\%$ more heavy hitters in comparison to the same dimension limit of 1 million when using 4 iterations. For these experiments we set the $\falseratio = 0.5$. However the empirical $\falseratio$ is $0.42$, $0.46$, $0.33$, and $0.42$ for $1M$ + 1 extra iteration, $1M$, $10M$, and $20M$ dimension limit, respectively.

\longversion{We explored other constraints effect on the utility of the algorithm in Appendix~\ref{appendix:tuning}. One other important constraint is the number of payloads a system can receive in a single iteration. In order to limit the number of payloads, we use Poisson sampling on the devices at each iteration, so each device tosses a coin and decides to participate with probability $p$. This reduces the number of devices participating, and allows us to take advantage of privacy amplification by sampling. For the evaluations, we use the privacy amplification by sampling for R\'enyi DP bounds due to~\cite{{zhu2019poission}}. }
\begin{figure*}
\centering
\begin{subfigure}[b]{0.3\textwidth}
    \centering
    \includegraphics[width = \textwidth]{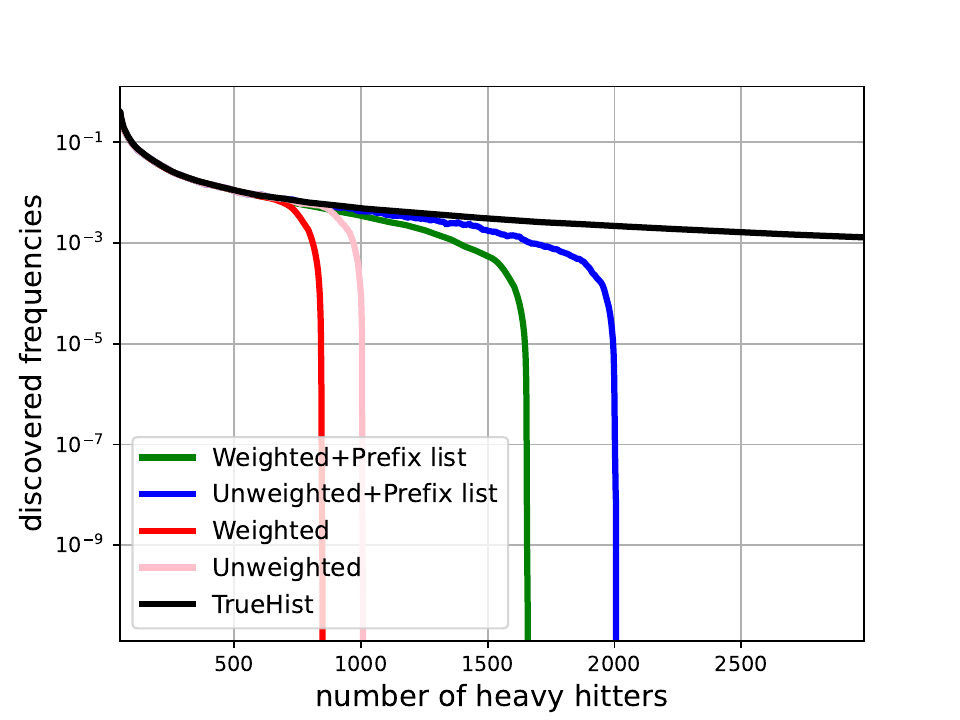}
    
    \caption{Discovered frequencies for different data selection schemes in $\ouralgorithm$ }
    \label{fig:weightfreqsamp}
\end{subfigure}
\hfill
\begin{subfigure}[b]{0.3\textwidth}
    \centering
    \includegraphics[width = \textwidth]{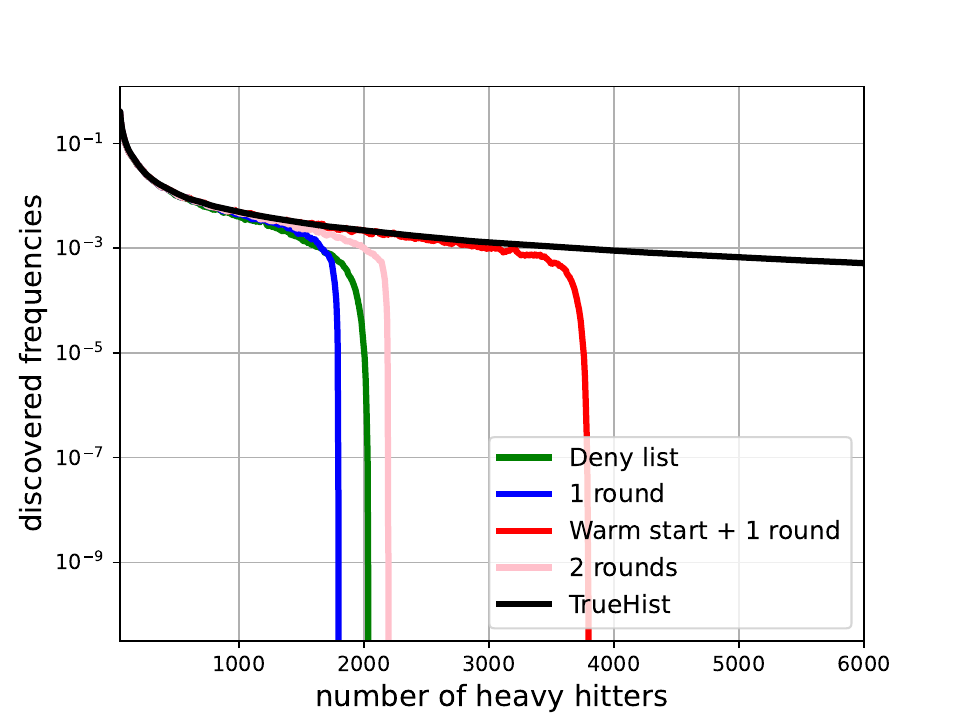}
   
    \caption{Discovered frequencies when adding deny list in different settings to $\ouralgorithm$}
    \label{fig:weightfreqdeny}
\end{subfigure}
\hfill
\begin{subfigure}[b]{0.3\textwidth}
    \centering
    \includegraphics[width = \textwidth]{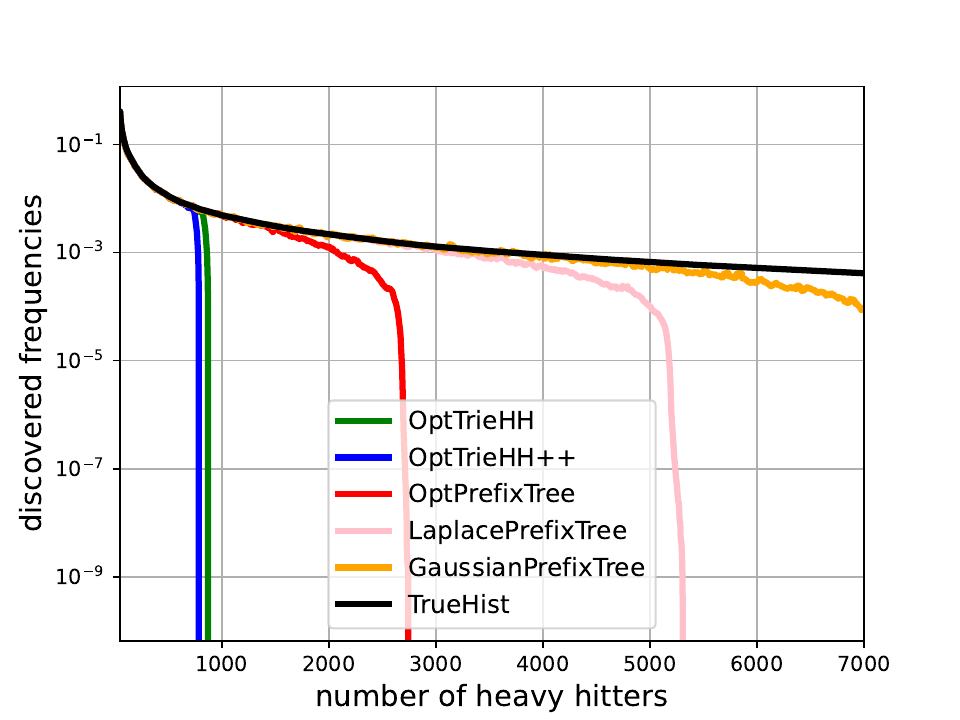}
    
    \caption{Discovered frequencies comparison between $\ouralgorithm$ and other methods}
    \label{fig:compfreqmulti}
\end{subfigure}

\caption{Experimental results in the multiple data points per \device setting with $\epsagg=1$, $\delta=10^{-6}$}

\end{figure*}
\subsection{Data Selection for Multiple Data Points per \Device}

Now we will explore the setting where each \device has multiple data points. \longversion{In these experiments we will set $\epsagg=1$ and $\delta = 10^{-6}$.}

\paragraph{Effect of Data Selection}
First, we evaluate the effect of different data selection schemes, specifically focusing on weighted vs unweighted data selection. In order to highlight the benefit of conditioning on the prefix list during data selection, we also compare against the version of these schemes that do not take the prefix list into account when selecting a data point. We will refer to the version where the selected data point is conditioned to belong in the prefix list as weighted selection + prefix list and unweighted selection + prefix list, and the versions where the selected data point is not forced to be in the prefix list as weighted selection and unweighted selection.
\longversion{In this experiment we used $\numiters = 4$.} Fig.~\ref{fig:weightfreqsamp} shows the experimental results using the different data selection schemes. Perhaps surprisingly, unweighted sampling outperforms weighted sampling in both the with and without prefix list experiments. One explanation for this observation is that in unweighted selection, moderately frequent words are given more opportunity to participate to the final output, while in weighted selection, the most frequent words are selected by all the users. That is, perhaps unweighted selection allows us to explore further into the tails of the distributions. As expected, conditioning on the prefix list has a significant impact. For the rest of this paper, we use unweighted sampling conditioned on the prefix list for on-device data selection. For these experiments we set the $\falseratio = 0.5$. However the empirical $\falseratio$ is $0.30$, $0.35$, $0.35$, and $0.38$ for weighted selection + prefix list, unweighted selection + prefix list, weighted selection, and unweighted selection respectively. For more evaluations please refer to Fig.~\ref{fig:selection}.

\paragraph{Effect of adding a deny list
} In practice, it is possible that an analyst has prior knowledge about some of the heavy hitters before the start of the algorithm. In such cases, we can take advantage of a deny list which includes this prior knowledge. When sending query to the \devicesnospace, we can also send the deny list and ask the \devices to exclude these already discovered data points during data selection. We explore two ways of obtaining a deny list, $\denylist$. In the first case, the deny list comes from an auxiliary public data source (which is not protected with DP), we will refer this as a \texttt{warm start}. The second is when we run the algorithm twice, with the heavy hitters from the first round forming the deny list of the second round (referred to as \texttt{2 rounds}). In \texttt{2 rounds} case, we need to account for the privacy budget of \emph{both} rounds,\longversion{ which leads to more noise per iteration}.
In Fig.~\ref{fig:weightfreqdeny} we compare the different configurations. The warm start is initiated with the top $2000$ popular words from Twitter Sentiment140 dataset~\cite{go2009twitter}. We also present the utility of the warm start $\denylist$ on its own (denoted by \texttt{deny list} line), showing the difference between the distribution of the two datasets.\longversion{ For comparison, we also include the standard algorithm run for a single round, without a deny list (denoted by \texttt{1 round} line).}  We further show the benefit of using deny list in one round of algorithm execution. Finally, we first execute one round of algorithm without a warm start and form a deny list out of discovered prefixes, which is then used in a second round. Fig.~\ref{fig:weightfreqdeny} shows that adding a deny list significantly improves performance. Adding warm start leads to discovering $2.1\times$ more heavy hitters. The 2 round algorithm increases the number of discovered heavy hitters by $1.22\times$. For twitter sentiment140 dataset only $0.006$ of the data points in this dataset do not belong to Reddit dataset. For the rest of configurations, we set the $\falseratio = 0.5$.  For \texttt{1 round} the empirical $\falseratio$ is $0.45$ while for \texttt{1 round + warm start} it's $0.37$. For \texttt{2 rounds} of algorithm $0.39$ of discovered bins are false positives. For more evaluations regarding deny list, please refer to Fig.~\ref{fig:deny}. 

\paragraph{Comparison to prior works} 

We compare with $\triehhg$ ($\triehh$) and $\triehhp$($\opttriehhp$), Central Laplace ($\laplacetree$), and Central Gaussian ($\gaussiantree$) with the optimisation that we use our adaptive segmentation to determine the segment length at each round, in the setting where each \device has multiple data points. For all of the algorithms we use the unweighted data selection, which performed the best in our previous experiments. To have a fair comparison, the same binary encoding is used for all of the models (5 bits per character, not Huffman encoding) and we used 12 iterations (1 character per iteration) that shows the best performance for this encoding for all the methods.

In $\triehh$ and $\opttriehhp$, for $\epsagg=1$ and $\delta = 10^{-6}$, we set the threshold for the number of reports that needs to be received for a word to be a part of the prefix list (denoted by $\theta$ in their paper) to $10$. Accordingly we set the sampling rate based for $\triehhg$ based on Corollary 1 in \cite{zhu2020federated} and for $\triehhp$ based on Lemma 3 in~\cite{cormode2022sample}. Our analysis in Fig.~\ref{fig:compfreqmulti} indicates $\ouralgorithm$ is able to discover $3.2$ times more heavy hitters for the same number of iterations and same dimension constraint. One explanation for this performance difference is that $\triehh$ and $\opttriehhp$ use sampling and thresholding to achieve the aggregated privacy guarantee, without adding any local differential privacy noise. When we set the threshold ($\theta$) to 10, the sampling rate is $0.0079$ and $0.0071$ for $\triehhg$ and $\triehhp$ respectively. Hence, the low sampling rates required to achieve the privacy guarantee results in sampling error in the distribution that is larger than the noise injected by our mechanism. In Appendix~\ref{appendix:TrieHH} and Appendix~\ref{appendix:TrieHH++} we explore the effect of different numbers of iterations in the utility of $\triehhg$ and $\triehhp$ for both single and multiple data points setting.

We further compare our method with a state-of-art central differential privacy methods. The assumption here is that a trusted curator is able to collect the \devicenospace's data and add central noise. We used two types of central noise for our analysis. For providing the central DP privacy guarantee we add Gaussian or Laplace noise to each dimension of the histogram~\cite{dwork2014algorithmic, dwork2006differential}. As with our previous experiments we compute the privacy guarantee of the composition over multiple rounds using the advanced composition theorem~\cite{DRV, Kairouz:2017} and the composition bound in terms of R\'enyi differential privacy~\cite{DLDP, mironov2017renyi}. We compute the R\'enyi privacy guarantees of both noise addition methods using the bounds provided in Table 2 of ~\cite{mironov2017renyi}. We finally select the noise power using the tighter bound between these two. As shown in Fig.~\ref{fig:compfreqmulti} both of the central DP scenarios outperform $\ouralgorithm$. While $\gaussiantree$ can be implemented in our model (by distributing the Gaussian noise among the participating devices), it suffers from a significantly worse local privacy guarantee.

\section{Conclusion}\label{sec:conclusion}

In this work we shed light on the importance of adaptive segmentation and intelligent data selection in heavy hitter detection algorithms. We conducted various experiments to find the optimum adaptive segmentation scheme based on the computation and communication constraints. In addition to comparing different data selection schemes, we demonstrated the benefit of using a prefix list and deny list for improving the utility of $\ouralgorithm$. Moreover, $\ouralgorithm$ is designed to satisfy both local and aggregated differential privacy. 

\newpage
\bibliography{main}
\bibliographystyle{apalike}

\newpage
\appendix
\onecolumn
\section{Notation table}\label{appen:not-table}

\begin{table}[h]
\centering

\begin{tabular}{|c|c|}

\hline
\textbf{Notation} & \textbf{Definition} \\ \hline
$\numusers$ & Number of users \\ \hline
$\datap_{i,j}$ & $j^{th}$ Datapoint of user $i$ \\ \hline
$\ndp_i$ & Number of datapoints of user $i$ \\ \hline
$\userdist$ & True underlying distribution of users \\ \hline
$\worddom$ & Datapoint domain \\ \hline
$\alphadom$ & Alphabet domain \\ \hline
$\worduniv$ & Universe of all possible datapoints \\ \hline
$\numiters$ & Number of unknown dictionary rounds \\ \hline
$\seglent$ & Segment length in iteration $t$ \\ \hline
$\totlen$ & Fixed total length of all the words \\ \hline
$\prefixset_t$ & histogram in iteration $t$\\ \hline
$\prefixlist_t$ & Prefix list after iteration $t$\\ \hline
$\usercompdata_{i,t}$ & private data of user $i$ after iteration $t$\\ \hline
$\dpmult$ & threshold multiplier\\ \hline
$\complimit$ & Dimension limit\\ \hline
$\sigma$ & standard deviation of the noise\\ \hline
$\falseratio$ & Ratio of number of the false positives to total discovered\\ \hline
$\freqest(x)$ & estimated frequency of data point $x$ \\ \hline

\end{tabular}

\caption{Notations}
\label{table:notations}
\end{table}
\section{Differential Privacy}\label{appendix:privacy}

In this work, we will consider an algorithm that satisfies two levels of privacy protection appropriate for federated learning; differential privacy in the aggregate model and differential privacy in the local model. For more details on a potential system for achieving these guarantees please see \cite{mcmillan2022private}. In the federated setting where users may have more than one data point, there are two main choices for the granularity of the privacy guarantee: \devicenospace-level DP and event (or data point)-level DP. We will focus on the stronger of these two guarantees, \devicenospace-level DP, which protects against a user changing all of the data points associated to them.
We will introduce these two types of privacy guarantees in this section. Throughout the remainder of this section, when we refer to a user's data point, we are referring to their set of data points.

\subsection{Aggregate Differential Privacy}

In the aggregate model of differential privacy, we assume the existence of an aggregation protocol that sums the local reports before they are released to the analyst. The analyst still interacts with the clients in a federated manner to perform the algorithm, but the aggregation protocol guarantees that the analyst does not receive anything about the local reports \emph{except their sum}. The aggregate model of DP is a derivative of the more general and more common than aggregate DP shuffle model of differential privacy introduced in \cite{ErlingssonFMRTT19, CheuSUZZ19}. 

\begin{definition}\label{defnaggregateDP1}
A single round algorithm $\A$ is $(\eps, \delta)$-DP in the aggregate model if the output of the aggregation protocol on two datasets that differ on the data of a single individual are close. Formally, an algorithm $\A:\cD^n\to\cZ$ is $(\eps, \delta)$-DP in the aggregate model if the following conditions both hold:
\begin{itemize}
\item it can be written as $\A(d^{(1)}, \cdots, d^{(n)})=\phi(\aggregator(f(d^{(1)}), \cdots, f(d^{n)}))$ where $f:\cD\to\cZ$ is a randomized function that transforms that data, $\aggregator$ is an aggregation protocol, and $\phi:\cY^n\to\cZ$ is some post-processing of the aggregated report
\item for any pair of datasets $D$ and $D'$ that differ on the data of a single individual, and any event $E$ in the output space,
\[
\Pr(\A(D)\in E)\le e^{\eps} \Pr(\A(D')\in E)+\delta.
\]
\end{itemize}
Note that the post-processing function takes the aggregation as its input and does not have access to the individual reports.
\end{definition}
When $\delta>0$, we call this approximate DP.
When each user uses a local randomizer (i.e. the functions $f$ in Definition~\ref{defnaggregateDP1} are local randomizers), the privacy guarantee in the aggregation model can be bounded by a quantity that is a function of both $\eps_0$, the privacy guarantee in the local model, and $n$, the number of users that participate in the aggregation protocol~\cite{ErlingssonFMRTT19, CheuSUZZ19}. As the number of users increases, the privacy guarantee on the output of the aggregation protocol gets stronger; essentially each user gets ``lost in the crowd''. In this work, we will bound aggregate DP guarantee by the numerical bound on privacy amplification by shuffling due to \cite{feldman2023stronger}, who provide bounds for both approximate DP and a related privacy notion called R\'enyi DP. 

A multi-round algorithm $\A$ is $(\eps,\delta)$-DP in the aggregate model if it is the composition of single round algorithms which are DP in the aggregate model, and the total privacy loss of $\A$ is $(\eps,\delta)$-DP.
One can formulate a version of Definition~\ref{defnaggregateDP1} specifically for multi-round algorithms, for a more in-depth discussion see \cite{Jain2021ThePO}. There are a number of standard theorems for analysing the privacy guarantee of composing multiple differentially private algorithms~\cite{DMNS06, DRV}. When the number of iterations is small, the advanced composition theorem~\cite{DRV, Kairouz:2017} provides a tight analysis. When the number of iterations is large, a tighter analysis is obtained by computing the composition bound in terms of R\'enyi differential privacy~\cite{DLDP, mironov2017renyi} then converting this R\'enyi bound into an $(\epsilon, \delta)$-DP bound~\cite{canonne2020discrete}. In our experiments, we compute the composed privacy guarantee using both of these methods, then select the tighter bound.

Given an expected number of users, the number of iterations, and the desired aggregate privacy guarantee, we can use binary search to approximate the largest per iteration local epsilon that will achieve the given aggregate privacy guarantee. This algorithm is given in Algorithm~\ref{alg:privacy} where RenyiShuffleAnalysis computes the R\'enyi privacy guarantee for amplification by shuffling, Composition uses the composition theorem for R\'enyi DP, Conversion converts the R\'enyi DP guarantees to approximate DP guarantees and BinarySearch makes the decision on whether to increase or decrease $\epslocal'$. Since the aggregate privacy guarantee is our primary privacy guarantee, we do not put an upper bound on our local epsilon. Table~\ref{tab:eps} demonstrates how the local epsilon increases with the number of iterations, and the desired aggregate privacy guarantee. Table~\ref{tab:eps} shows different values of $\epslocal$ and $\epsagg$ depending on the number of iterations for $\numusers = 1.6 \times 10^6$ devices (number of users in the Reddit data set used for our experiments). 

\begin{algorithm}
\caption{Privacy Analysis}\label{alg:privacy}
\begin{algorithmic}[1]
\STATE \textbf{Input}: $\epsilon_{agg},\delta$: Aggregate privacy budget, $\numiters:$ number of iterations, $\numusers$: number of \devices, $\alpha$: pre-defined set of Renyi parameter, $E$: binary search error tolerance
\STATE \textbf{Output}: $\epslocal$: local privacy budget of each \device in each iteration
\STATE $\epslocal' \gets \text{Initialization}$
\WHILE {$|\epsilon_{agg}-\epsilon'_{agg}| \le E$}

\STATE $\epsilon'_r \gets \text{RenyiShuffleAnalysis}(\epsilon'_l, \delta, \numiters, \numusers, \alpha)$ \textit{{\color{darkgreen} // Theorem 3.2 of ~\cite{feldman2023stronger}}}
\STATE $\epsilon'_{composition} \gets \text{Composition}(\epsilon'_r, \numiters)$ \textit{{\color{darkgreen} //  $\epsilon'_{composition} = \numiters \times \epsilon'_r$ }}
\STATE $\epsilon'_{agg} \gets \text{Conversion}(\epsilon'_{composition}, \delta, \alpha)$ \textit{{\color{darkgreen} //  Proposition 12 of ~\cite{canonne2020discrete}}}
\STATE $\epsilon'_l \gets \text{BinarySearch}(\epsilon_{agg}, \epsilon'_{agg}, \epslocal')$
\ENDWHILE
\STATE $\epsilon_l \gets \epsilon'_l$
\STATE \textbf{return} $\epsilon_l$
\end{algorithmic}
\end{algorithm}

\begin{table}[h]
\centering
\resizebox{\columnwidth}{!}{%
\begin{tabular}{|c|cccccc|cccccc|cccccc|}
\hline
$\epsilon_{agg}$    & \multicolumn{6}{c|}{$0.25$}                                                                                                         & \multicolumn{6}{c|}{$0.5$}                                                                                                         & \multicolumn{6}{c|}{$1$}                                                                                                         \\ \hline
\textbf{$\numiters$}      & \multicolumn{1}{c|}{1} & \multicolumn{1}{c|}{2} & \multicolumn{1}{c|}{3} & \multicolumn{1}{c|}{4} & \multicolumn{1}{c|}{5} & 6 & \multicolumn{1}{c|}{1} & \multicolumn{1}{c|}{2} & \multicolumn{1}{c|}{3} & \multicolumn{1}{c|}{4} & \multicolumn{1}{c|}{5} & 6 & \multicolumn{1}{c|}{1} & \multicolumn{1}{c|}{2} & \multicolumn{1}{c|}{3} & \multicolumn{1}{c|}{4} & \multicolumn{1}{c|}{5} & 6 \\ \hline
\textbf{$\epsilon_l$} & \multicolumn{1}{c|}{6.36}  & \multicolumn{1}{c|}{6.05}  & \multicolumn{1}{c|}{5.79}  & \multicolumn{1}{c|}{5.63}  & \multicolumn{1}{c|}{5.35}  & 5.31  & \multicolumn{1}{c|}{7.18}  & \multicolumn{1}{c|}{6.96}  & \multicolumn{1}{c|}{6.73}  & \multicolumn{1}{c|}{6.48}  & \multicolumn{1}{c|}{6.33}  & 6.26  & \multicolumn{1}{c|}{8.03}  & \multicolumn{1}{c|}{7.73}  & \multicolumn{1}{c|}{7.58}  & \multicolumn{1}{c|}{7.39}  & \multicolumn{1}{c|}{7.03}  & 7.018  \\ \hline
\end{tabular}
}
    \caption{Choices of $\eps_{agg}$, $\numiters$ and $\eps_l$ }
    \label{tab:eps}
\end{table}

\section{PI-RAPPOR}\label{appen:pi-rappor}

In this section, we describe the PI-RAPPOR local randomizer that may be used as the local randomizer ($\compressor$) and two algorithms ($\decompressor$) we can use to get the private frequency estimates each element queried (usually the data domain in each round $\prefixlist_{t-1}\times \alphadom^{\seglent}$). We also discuss the computation and communication costs for the device side and both the server side algorithms, provide recommendations for speeding up the algorithms by a factor of $e^\epslocal + 1$, and provide guidelines for practitioners on how to choose one amongst the two frequency estimation algorithms based on the computation costs. 

As defined in \cite{feldman2021lossless}, we define two constants $\alpha_0$ and $\alpha_1$ which we set suitably to satisfy deletion $\epslocal$ - DP ($\alpha_0 = 1 - \alpha_1 = \frac{1}{e^\epslocal + 1}$) and replacement $\epslocal$ - DP ($\alpha_0 = \frac{1}{e^\epslocal + 1}, \alpha_1 = \frac{1}{2}$), respectively. Let $q$ be a prime power so that $\alpha_0 q$ is an integer (the case when $\alpha_0 q$ is not an integer incurs a small additional error as described in Lemma 4.7 of \cite{feldman2021lossless}). We let $\smallfield$ denote the $\alpha_0 q$ smallest elements of the field and let $\bool{z}$ denote the indicator of the event $z \in \smallfield$. Algorithm 3 of \cite{feldman2021lossless} gives the PI-RAPPOR local randomizer algorithm which can be used as $\compressor$ and Algorithms 4 and 5 of \cite{feldman2021lossless} are two frequency estimation algorithms, which use the outputs of $\compressor$ applied to a datapoint from all users and estimate the frequency of elements in the data domain. The frequency estimate output of either of these algorithms of an element $w$ is a random variable $\freqest[w]$ as defined below:
\begin{equation}
    \tilde{f}[w] = \frac{\binomial{f[w]}{\alpha_1} + \binomial{n - f[w]}{\alpha_0} - n\alpha_0}{\alpha_1 - \alpha_0},
\end{equation}
where $f[w]$ is the true frequency of $w$ and $\binomial(k,\alpha)$ denotes a binomial random variable with parameters $k$ (number of experiments) and $\alpha$ (success probability).
 
As shown in Lemma 4.2 of \cite{feldman2021lossless}, $\tilde{f}$ is an unbiased estimator of $f$ with the minimum possible variance for locally private estimators. Algorithms 4 and 5 in \cite{feldman2021lossless} output the same estimate but only differ in terms of their computational complexity. Please See \cite{feldman2021lossless} for a discussion. While both the algorithms find the frequency of all elements in the domain, they can be easily modified to find the frequency of only subset of the elements in the domain and the computational complexity correspondingly depends on this subset size linearly.

\paragraph{Speedups in the decompressor for prefix based algorithms:}

When the data domain is a contiguous set of elements, we can 
precisely calculate the values of $w$ for which it evaluates to $\bool{v(w)} = 1$ and use it to speed up computation by a factor of roughly $\frac{1}{\alpha_0}$. For a given segment length $\seglen$, we choose $q$ to be the smallest prime power bigger than $K = \max\{e^\epslocal + 1,2^\seglen\}$. We also let $c = \lfloor \frac{q}{e^\epslocal + 1} \rfloor$ and denote $\{0,1,\dots,c\}$ by $[c]$. The data domain is of the form $\prefixlist \times \field{q}$ (mapped to $\field{q}^d$) where the elements of $\prefixlist$ are mapped to elements in the first $d - 1 = \lceil\log_q|\prefixlist|\rceil$ dimensions and the last dimension is assigned to all possible completions for a given prefix of length $\seglen$. 

Both algorithms 4 and 5 of \cite{feldman2021lossless} calculate $\bool{v^i(w)}$ where $v^i(w)= v^i_0 + \sum_{j \in [d]} v^i_jw_j$ for all $w$ in the data domain, which in our case is $\prefixlist \times \field{q}$. Instead of calculating the product $v^i(w) = v^i_0 + \sum_{j \in [d]} v^i_jw_j$ for all $w \in \prefixlist \times \field{q}$, we calculate  $v^i_{-1}(h) = v^i_0 + \sum_{j \in [d - 1]} v^i_jh_j$ for all prefixes $h \in \prefixlist$. Then for each element (say $g$) in $[c]$, we calculate $(v^i_d)^{-1}(g - v^i_{-1}(h))$ to get back precisely the element $w_d \in \field{q}$ for which $\bool{v^i(w)}$ will evaluate to 1. The inverse operation can be computed using a lookup table that can be calculated ahead of time in $\log(q)$ time. Thus, instead of searching over all elements in the data domain $\prefixlist \times \field{q}$ in constant time, we search over all possible cutoffs and prefixes in $\log(q)$ time each making the effective computational complexity linear in $c|\prefixlist|\log(q)$ instead of linear in $|\prefixlist|q$, which is a speedup of size roughly $\frac{e^\epslocal + 1}{\epslocal}$.

\section{Proof of \Cref{prop:one-round}}\label{appendix:proof}

\begin{proof}[Proof of \Cref{prop:one-round}]
    In a single round, we query each element in the data domain of size $|\alphadom|^\totlen$. Thus using \Cref{ass:local-rand-perf}, with $\beta' = \frac{\beta}{|\alphadom|^\totlen}$ and using a union bound, we have that with probability $1 - \beta$, $\max_{d \in \worduniv} |F[d] - \freqest[d]| \leq C\sqrt{\frac{ne^\epslocal}{(e^\epslocal - 1)^2}\log(\frac{|\alphadom|^\totlen}{\beta})}$. 

    Now, we prove that this algorithm is $\lambda$-accurate (\Cref{def:lam-acc}) with $\lambda = C\sqrt{\frac{ne^\epslocal}{(e^\epslocal - 1)^2}\log(\frac{|\alphadom|^\totlen}{\beta})}$. Let $d \in \alphadom^\totlen$ with $\freqest[d] \geq \tau\sigma + \lambda$. Then, $f[d] \geq \tau\sigma $ and hence $d \in \prefixlist_1$. Next, let $d \in \alphadom^\totlen$ such that $f[d] < \tau\sigma - \lambda$, then $f[d] < \tau\sigma$ and we have $d \notin \prefixlist_1$. This shows that $\prefixlist_1$ is $\lambda$-accurate with $\lambda = C\sqrt{\frac{ne^\epslocal}{(e^\epslocal - 1)^2}\log(\frac{|\alphadom|^\totlen}{\beta})}$.
    
\end{proof}

\section{Adaptive Segmentation Exploration}\label{appendix:tuning}

Here, in addition to showing discovered frequencies and discovered counts, we show another metric which we refer to as utility loss.

Let $H = (x_1,x_2,\dots,x_{|H|})$ denote the set of heavy hitters output by an algorithm ordered by the empirical global frequency distribution $\empuserdist$, and let $x_i^*$ denote the $i^{\rm th}$ most frequent element according to the global empirical distribution $\empuserdist$, i.e. the true $i^{\rm th}$ heavy hitter. Then, we evaluate an algorithm with output $H$ by how close the total mass of $H$ is to the total mass of the true top $|H|$ heavy hitters. 

\textit{Utility Loss:} Define the weight ratio as $\weightratio{(H)} = \frac{\sum_{x \in H} \empuserdist(x)}{\sum_{i \in [|H|]} \empuserdist(x_i^*)}$, i.e. the ratio of total probability mass of private heavy hitters $H$ over the probability mass of the actual top $|H|$ heavy hitters. The goal is to maximize the $\weightratio$ to minimize the loss (1-\weightratio).

One other potential metric for evaluating these models is to use precision and recall or different versions of their combination for instance F1-Score. However, these metrics do not take into the account the frequency of discovered items. Meaning that if any iterative algorithm discovers the most frequent heavy hitters, but some of them are not in the actual most frequent heavy hitters because of a small frequency difference, precision/recall metrics are not able to capture that. In other words, they capture those as a miss which is not fair to these algorithms.

For marginal figures, to aid with visualisation in our plots, for a window size $W=50$ and a given heavy hitter set $H$, we plot for each $i$, the sum of the probabilities (according to $\empuserdist$) of the heavy hitters in the sliding window $(x_{i - W},x_{i-W + 1},\dots, x_i)$. We also plot a true histogram line representing $(x^*_{i - W},x^*_{i-W + 1},\dots, x^*_i)$ (TruHist line) as a reference of what true histogram looks like. 

\subsection{Adaptive Segmentation for Single Data Point}
In this section, we explore the effect of different parameters in the utility of $\ouralgorithm$. For simplicity in this section, we assume each \device has a single datapoint.

\paragraph{False Positives Ratio}

In these experiments we investigate the effect of different $\falseratio$ parameters on the utility of the final model. False positives ratio can be defined for each application. Depending on how sensitive the application is, $\falseratio$ determines ratio of number of expected false positives to the total discovered that we keep in each iteration. Please note that this parameter is application dependent. For the applications that are more tolerant to false positives this ratio can be higher. In this part we set the $\complimit = 10^7$ and $\ouralgorithm$ finishes in 4 iterations.

\begin{figure}[htb]
     \centering

        \begin{subfigure}{2in}
         \centering
         \includegraphics[width=2in]{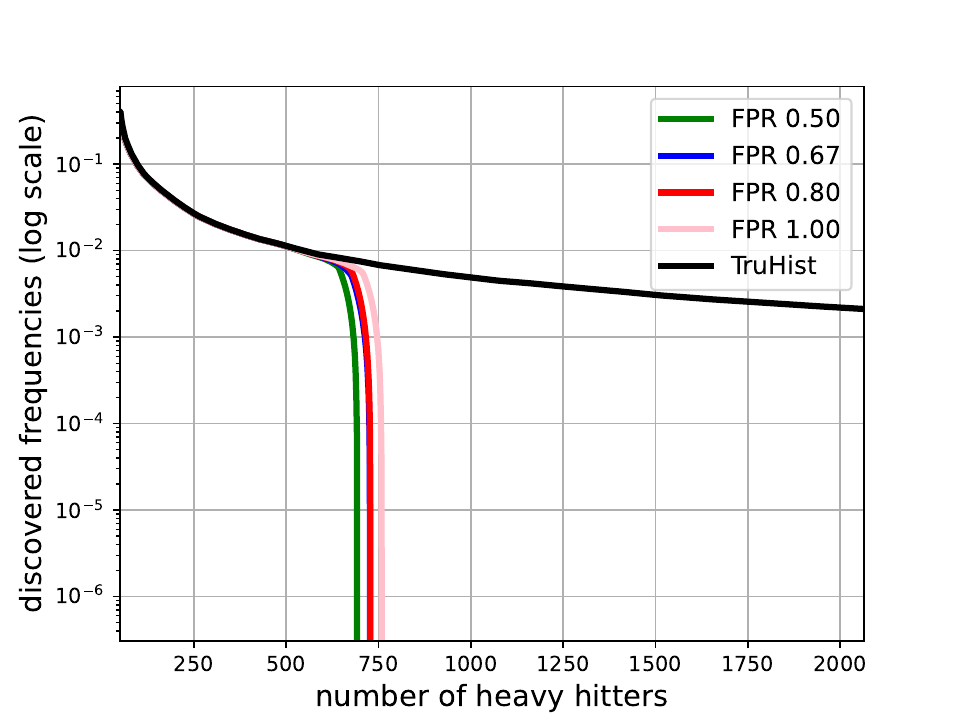}
     \caption{Marginal discovered frequencies}
    \label{fig:trueratiofreq}
     \end{subfigure}
     \hfill
     \begin{subfigure}{2in}
         \centering
         \includegraphics[width=2in]{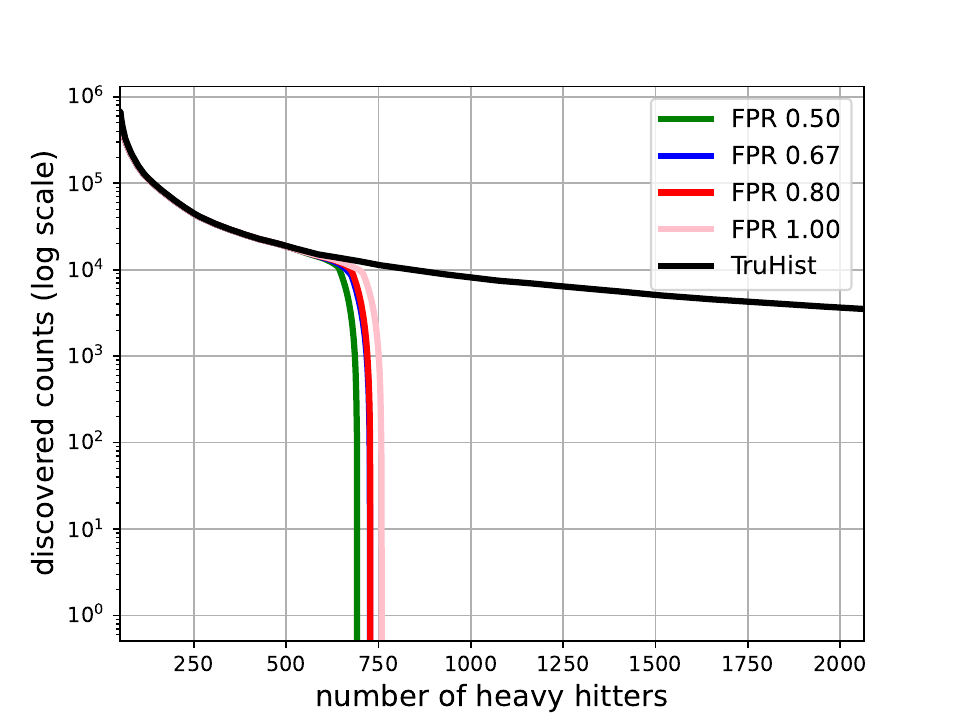}
            \caption{Marginal discovered counts}
        \label{fig:trueratiocount}
     \end{subfigure}
     \hfill
          \begin{subfigure}{2in}
         \centering
         \includegraphics[width=2in]{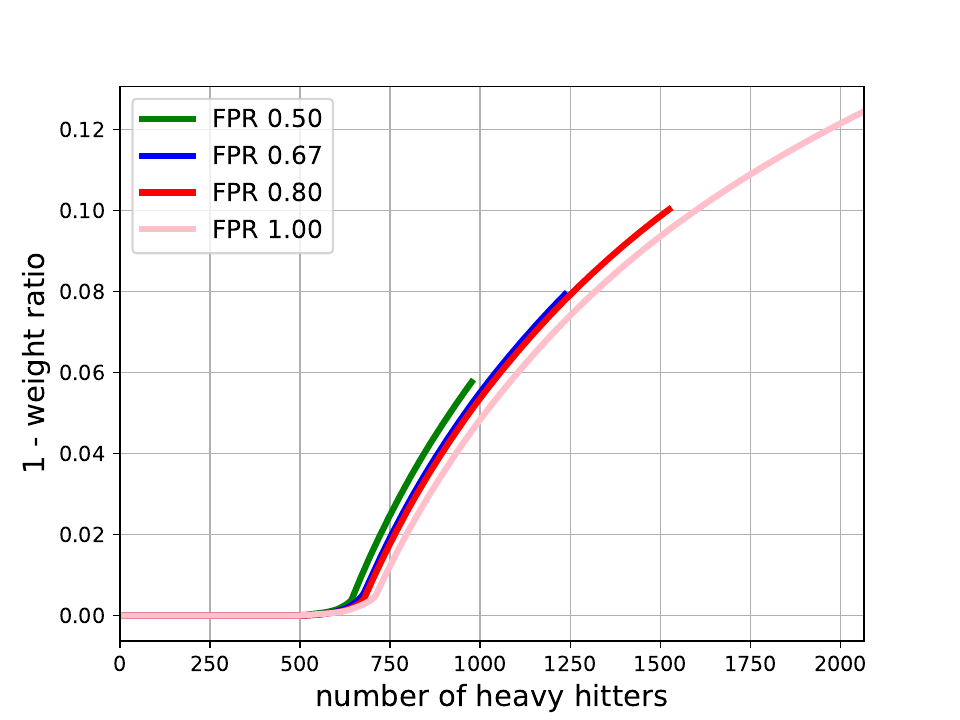}
            \caption{Total $\weightratio$ Loss}
        \label{fig:trueratiototal}
     \end{subfigure}

    \caption{Effect of the different $\falseratio$ parameters for single data point setting on $\ouralgorithm$ utility ($\epsagg=1$, $\delta=10^{-6}$ and $T=4$)}
 \end{figure}

As observed increasing the $\falseratio$ from 0.5 to 1 increases the number of heavy hitters detected slightly [660, 656, 678.0, 695] but increasing $\falseratio$ also increases the number of discovered false positives significantly [357, 636, 828, 1400]. Based on Figure~\ref{fig:trueratiototal}, although lower $\falseratio$ detects fewer true bins to ensure it includes fewer false positives, it detects the top most-frequent bins correctly as the loss value shown on $y$-axis is negligible. We also remark that $\falseratio=1$ does not imply that all bins are included since the threshold is based on the expected expected value of false positives for every confidence whereas the true number typically fluctuates around the expectation.

 \paragraph{Number of Devices Limitation}

\begin{figure}[htb]
     \centering

          \begin{subfigure}{2in}
         \centering
         \includegraphics[width=2in]{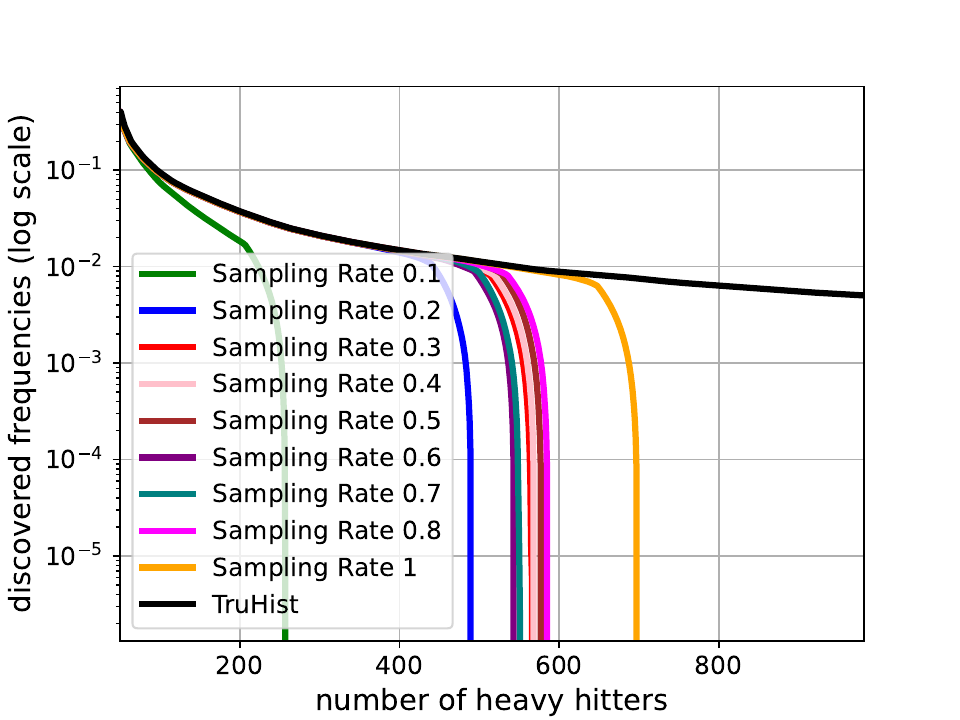}
     \caption{Marginal discovered frequencies}
    \label{fig:sampfreq}
     \end{subfigure}
     \hfill
     \begin{subfigure}{2in}
         \centering
         \includegraphics[width=2in]{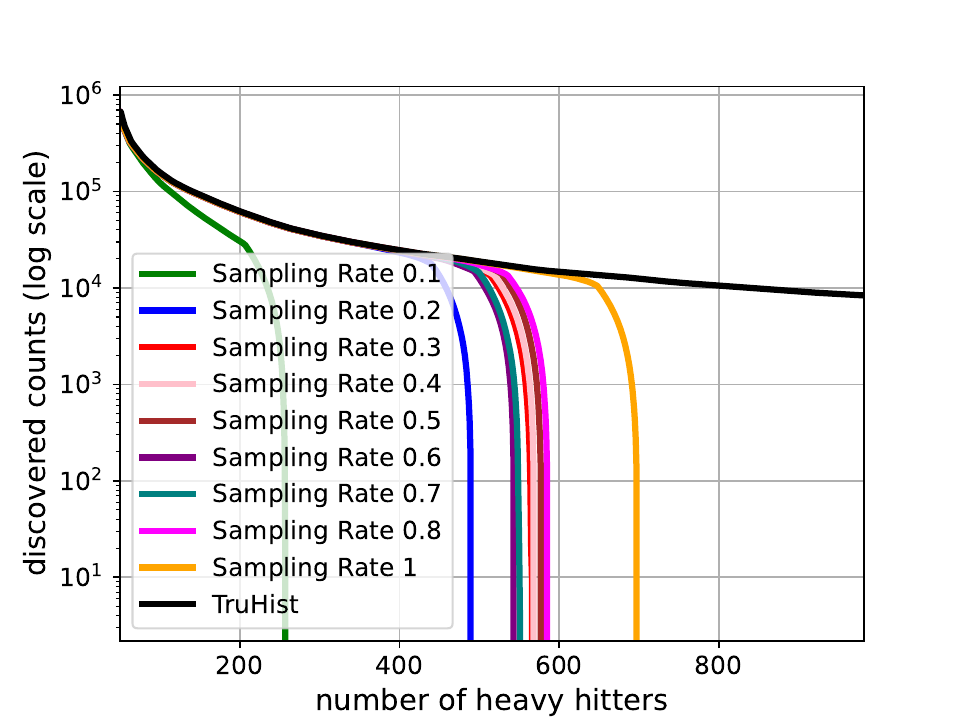}
     \caption{Marginal discovered counts}
    \label{fig:sampcount}
     \end{subfigure}
     \hfill
    \begin{subfigure}{2in}
         \centering
         \includegraphics[width=2in]{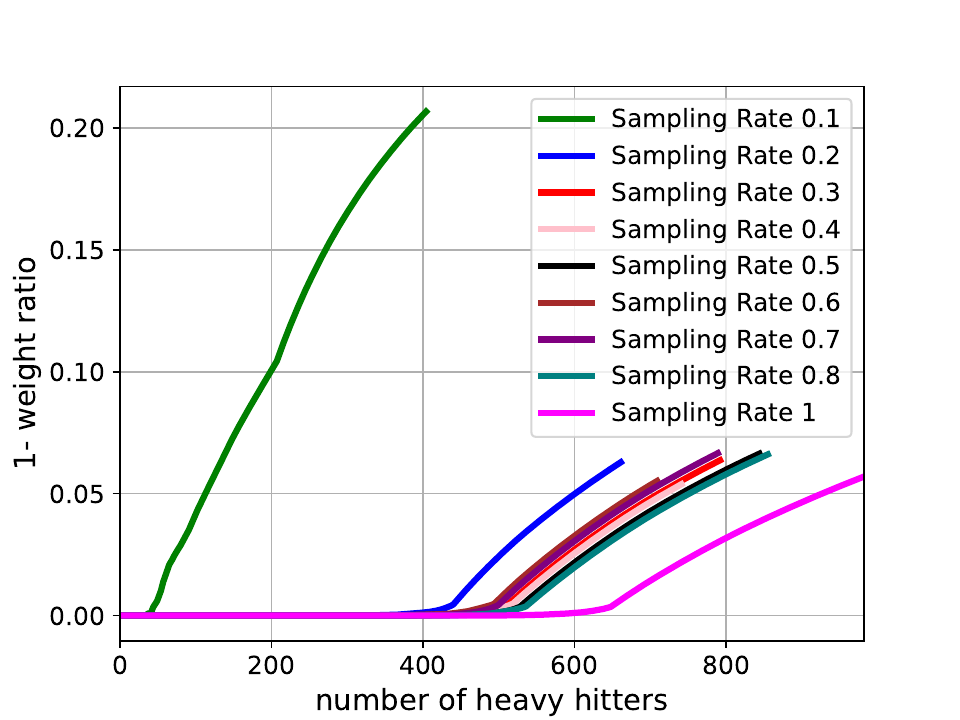}
     \caption{Total $\weightratio$ Loss}
    \label{fig:totalsamp}
     \end{subfigure}
     \caption{Effect of the different sampling rates for single data point setting on $\ouralgorithm$ utility ($\epsagg=1$, $\delta=10^{-6}$ and $T=4$)}   
 \end{figure}

 In this part we analyze the effect of having constraints on the number of \devices on the utility of the model. In a production-scale model with billions of \devices sending data, dimension can grow extremely large. One way to get around this issue is to use sampling. By sampling in each iteration only a sub-set of \devices receive the query and contribute to the algorithm. In this section we discuss the effect of having different sampling rates on the utility of the model. Each \device can participate in an iteration with a $\bernoulli\gamma$ where $\gamma$ is the sub-sampling rate. We use theorem 5 described in~\cite{zhu2019poission} for estimating the upper-bound of $\epscentral$. The effect of different sub-sampling rates on the value of $\epslocal$ when $\epscentral=1$ and $\numusers = 1.6*(10)^6$ is shown in Table~\ref{tab:samp}. As shown small sampling rate, increase the $\epslocal$ by adding to the randomness. However, as shown in the table, no sampling leads to higher $\epslocal$ in comparison to the moderate sampling rates $\epslocal$. The reason is when the number of \devices increases, the privacy guarantee on the output of the aggregation protocol gets stronger (``lost in the crowd''). Hence, for moderate sampling rates having less number of users cancels the benefit of using sampling. Figure~\ref{fig:totalsamp}, ~\ref{fig:sampcount}, ~\ref{fig:sampfreq} demonstrates the effect of different sampling rates on the utility of $\ouralgorithm$. In addition to $\epslocal$ difference of using different methods, having smaller number of \devices can affect the utility by eliminating some part of the distribution. Consequently, we avoid using sampling if the dimension constraint allows.

 \begin{table}[]
    \centering
    \resizebox{\columnwidth}{!}{%
    \begin{tabular}{|c|c|c|c|c|c|c|c|c|c|c|c|c|c|c|c|c|c|c|c|c|c|c|c|c|c|c|c|c|c|c|}
    \hline
        $\numiters$ & \multicolumn{10}{|c|}{$3$} & \multicolumn{10}{|c|}{$4$} & \multicolumn{10}{|c|}{$5$} \\ \hline
        sampling rate&$0.1$ & $0.2$ & $0.3$ & $0.4$ & $0.5$ & $0.6$ & $0.7$ & $0.8$ & $0.9$ & $1$&$0.1$ & $0.2$ & $0.3$ & $0.4$ & $0.5$ & $0.6$ & $0.7$ & $0.8$ & $0.9$ & $1$& $0.1$ & $0.2$ & $0.3$ & $0.4$ & $0.5$ & $0.6$ & $0.7$ & $0.8$ & $0.9$ & $1$ \\ \hline
        $\epslocal$& $8.42$ & $8.26$ & $8.19$ & $8.12$ & $8.1$ & $7.92$ & $7.81$ & $7.73$ & $7.62$ & $7.73$ & $8.32$ & $8.19$ & $8.12$ & $7.97$ & $7.87$ & $7.72$ & $7.59$ & $7.52$ & $7.4$ & $7.58$ & $8.27$ & $8.12$ & $8.1$ & $7.86$ & $7.6$ & $7.35$ & $7.19$ & $7.2$ & $7.02$ & $7.39$  \\ \hline
    \end{tabular}
    }
    \caption{Sampling rate effect on the $\epslocal$ for $\epsagg = 1$ and $\delta = 10^{-6}$}
    \label{tab:samp}
\end{table}

 \subsection{Multiple Data point Adaptive Segmentation}\label{appendix:multioptprefixtree}
 \textbf{Segmentation Size}

 In this analysis, we used binary encoding of the data. We set the dimension limitation to $\complimit=10^7$. We ran the experiment for the utilized setting in which we have unweighted sampling and prefix list. We used both weighted and unweighted metric. As demonstrated increasing the number of iterations from 3 to 12 improves the utility of the algorithm.

\begin{figure}[htb]
     \centering
          \begin{subfigure}{3in}
         \centering
         \includegraphics[width=3in]{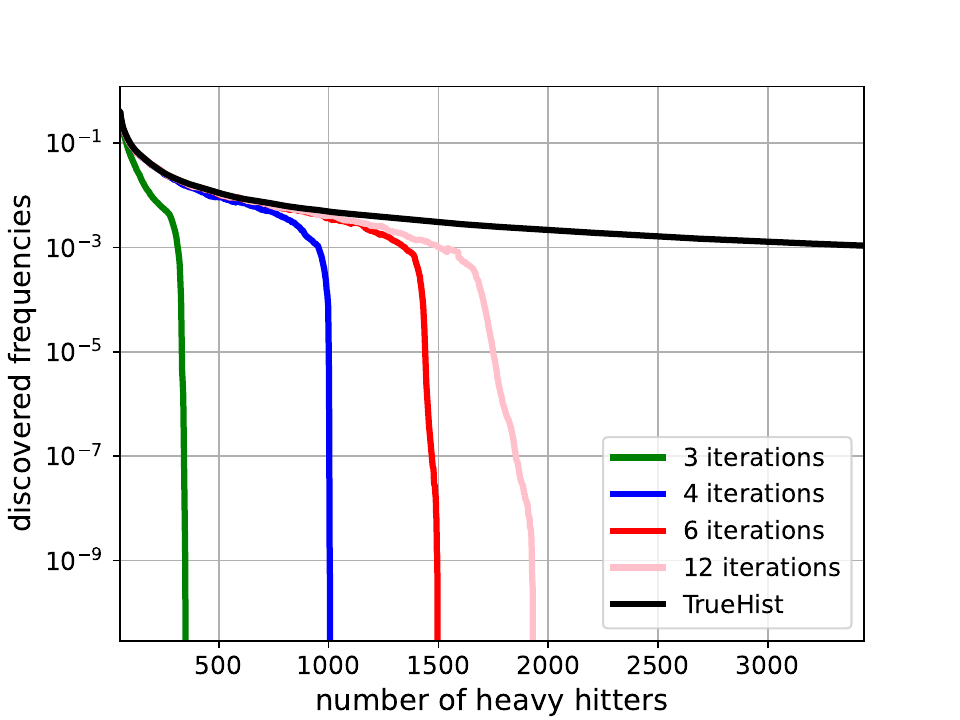}
     \caption{Marginal discovered frequencies (weighted metric)}
    \label{fig:LargeSegFreq}
     \end{subfigure}
     \hfill
     \begin{subfigure}{3in}
         \centering
         \includegraphics[width=3in]{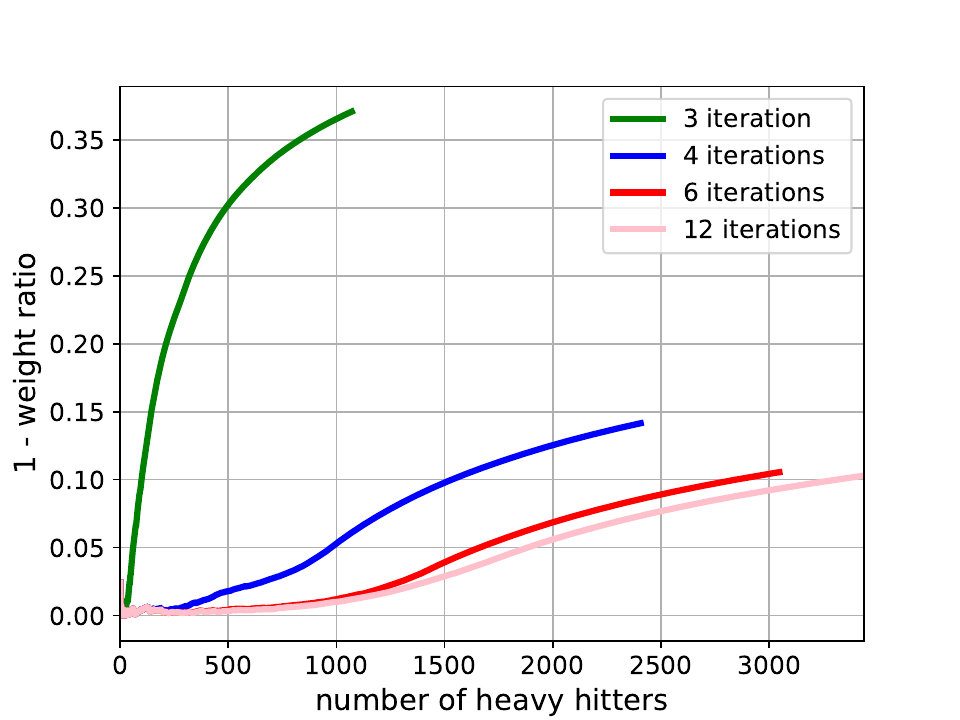}
     \caption{Total $\weightratio$ loss (weighted metric)}
    \label{fig:LargeSegTotalWeighted}
     \end{subfigure}

     \caption{Effect of the segment size for multiple data points setting on $\ouralgorithm$ utility ($\epsagg=1$, $\delta=10^{-6}$)}
 \end{figure}

 \begin{figure}[htb]
     \centering

     \begin{subfigure}{3in}
         \centering
         \includegraphics[width=3in]{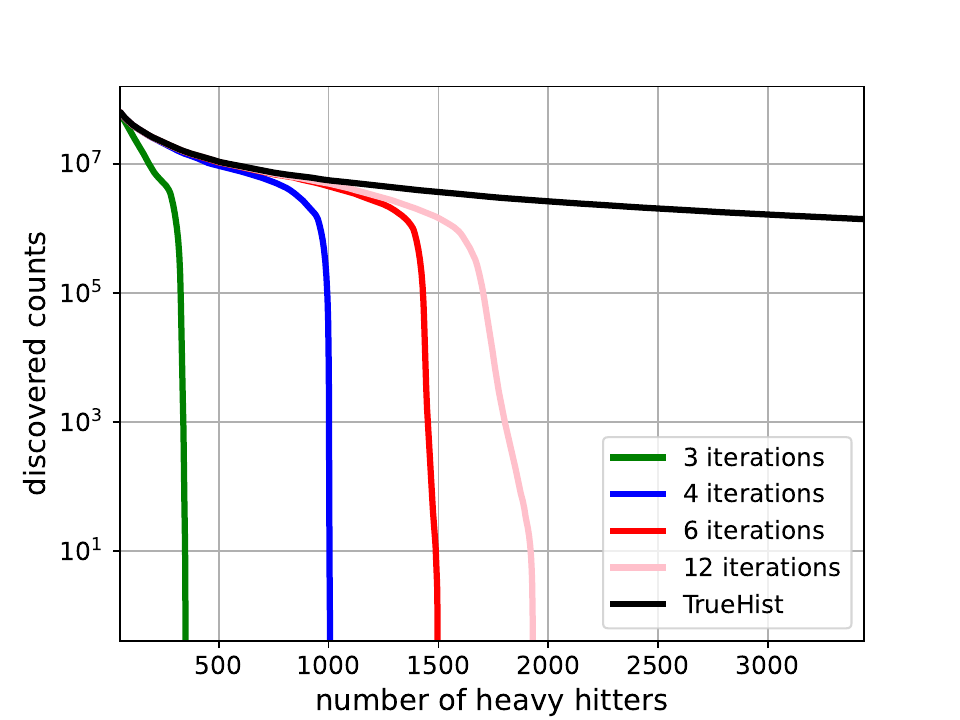}
     \caption{Marginal discovered counts (unweighted metric)}
    \label{fig:LargeSegCount}
     \end{subfigure}
     \hfill
          \begin{subfigure}{3in}
         \centering
         \includegraphics[width=3in]{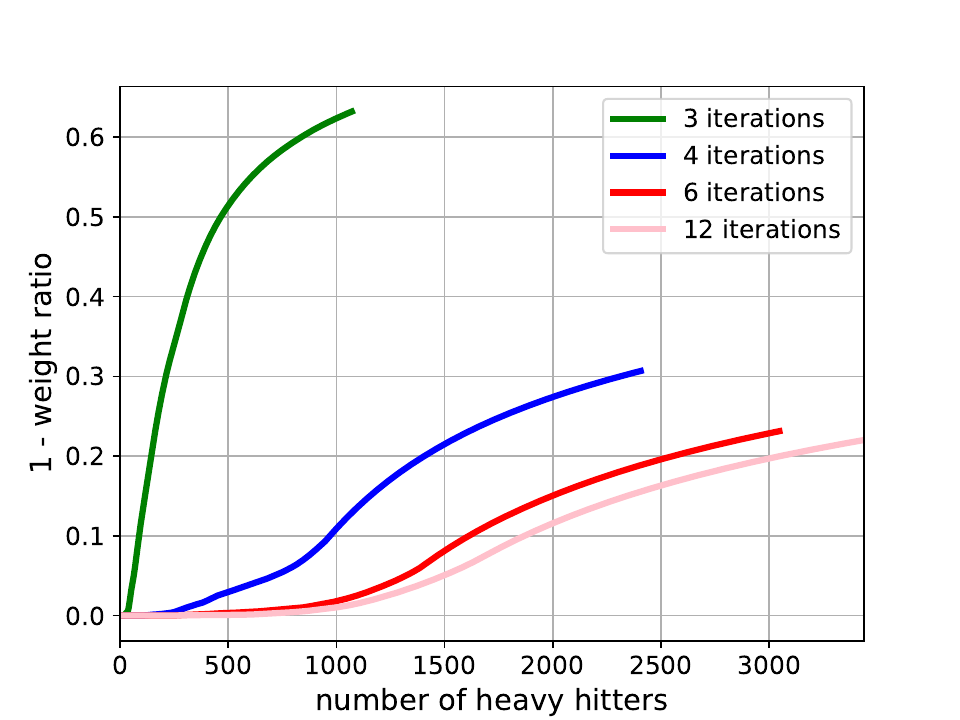}
     \caption{Total $\weightratio$ Loss (unweighted metric)}
    \label{fig:LargeSegtotal}
     \end{subfigure}
     \caption{Effect of the segment size for multiple data points setting on $\ouralgorithm$ utility ($\epsagg=1$, $\delta=10^{-6}$)}
 \end{figure}

\section{TrieHH}
\label{appendix:TrieHH}

\begin{table}[htb]
    \centering
    \resizebox{\columnwidth}{!}{%
    \begin{tabular}{|c|c|c|c|c|c|c|c|c|c|c|c|c|c|}
    \hline
        $\epsagg$ & \multicolumn{4}{|c|}{$1$} & \multicolumn{4}{|c|}{$0.5$} & \multicolumn{4}{|c|}{$0.25$} \\ \hline
        $T$ & $12$ & $6$  & $4$ & $3$  & $12$ & $6$  & $4$ & $3$& $12$ & $6$  & $4$ & $3$\\ \hline
         Sampling Rate  & $0.0079$ & $0.0153$ & $0.0221$& $0.0283$ & $0.0040$ & $0.0079$ & $0.0117$& $0.0153$   &$0.0020$ & $0.0040$ & $0.0060$& $0.0079$  \\ \hline
    \end{tabular}
    }
    \caption{Number of rounds effect on the sampling rate of $\triehhg$ ($\delta=10^{-6}$, $\numusers = 1.6\times10^6, \theta = 10$)}
    \label{tab:TrieHH}
\end{table}

\subsection{Single Data Point Setting for $\triehhg$}\label{appendix:single}
\begin{figure}[htb]
     \centering

     \begin{subfigure}{2in}
         \centering
         \includegraphics[width=2in]{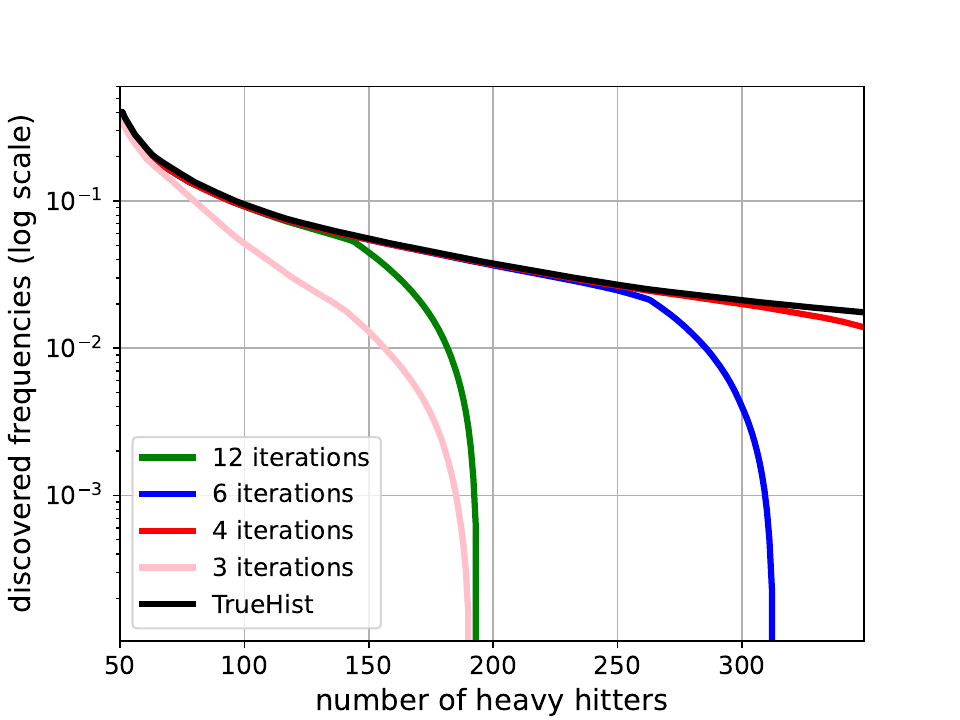}
    \caption{Marginal discovered frequencies}
    \label{fig:triehhfrequency}
     \end{subfigure}
     \hfill
          \begin{subfigure}{2in}
         \centering
         \includegraphics[width=2in]{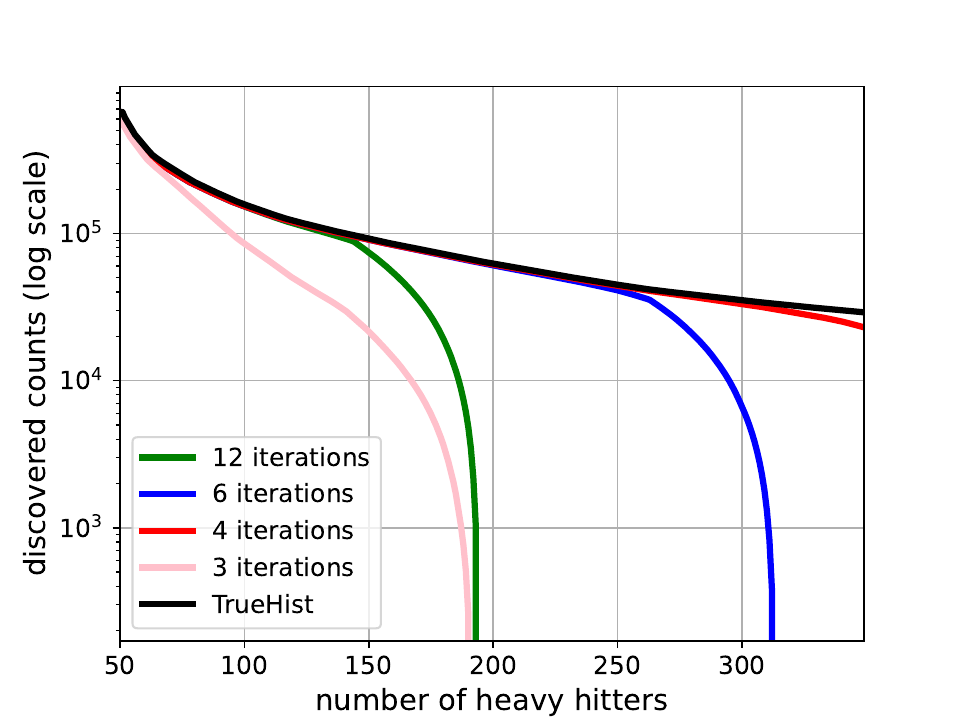}
    \caption{Marginal discovered counts}
    \label{fig:triehhcounts}
     \end{subfigure}
     \hfill
     \begin{subfigure}{2in}
         \centering
         \includegraphics[width=2in]{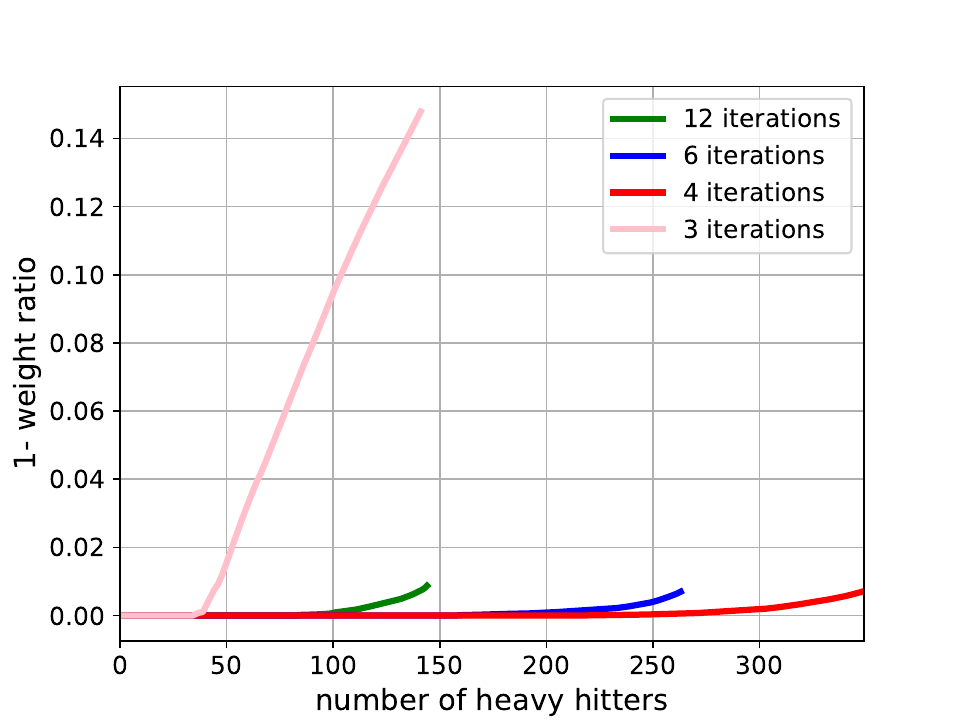}
    \caption{Total $\weightratio$ loss}
    \label{fig:triehhtotal}
     \end{subfigure}
    \caption{The effect of different number of iterations on $\triehhg$ for single data point setting ($\epsagg=1$, $\delta=10^{-6}$)}
 \end{figure}

In this section we discuss the effect of number of iterations on the utility of a single data point setting for $\triehhg$~\cite{zhu2020federated}. In Figure~\ref{fig:triehhfrequency}, we show the effect of different segmentation on the utility of the algorithm for a single data point per \device setting. For these experiments we used $\epsagg = 1$ and sampling rates are set based on table~\ref{tab:TrieHH}. To evaluate the effect of different segmentation in this part we used the $\complimit = 10^7$ on the dimension. The number of heavy hitters detected by $\triehhg$ algorithm, when the number of iterations are 12 (1 char), 6 (2 char), 4 (3 char), 3 (4 char) are $[142, 261, 355, 135]$. Initially having larger segments help with the algorithm since less number of iterations are required and consequently sampling rate becomes larger. However, by increasing the segment size to certain point, the utility drops. The reason is by enlarging the segment length the number of prefixes in each iteration reduces because of the dimension constraint. Hence, for the comparisons with $\ouralgorithm$ we used the best configurations which is having 4 iterations. 

\subsection{Multiple Data Points Setting for $\triehhg$}\label{appendix:multi}
We further analyze the multiple data points setting. To optimize the algorithm we took advantage of a prefix list for each iteration. \Device s send their data only if they find a match with a prefix in the prefix list. We also use an end character symbol to indicate the end of string. If end character symbol is observed in a prefix at the end of an iteration, the corresponding prefix will be excluded from the prefix list. Therefore, users can send other unfinished prefixes. Also, for our evaluation, we used binary encoding which uses $5$ bits to represent each character. 

\begin{figure}[htb]
     \centering
     \begin{subfigure}{3in}
         \centering
         \includegraphics[width=3in]{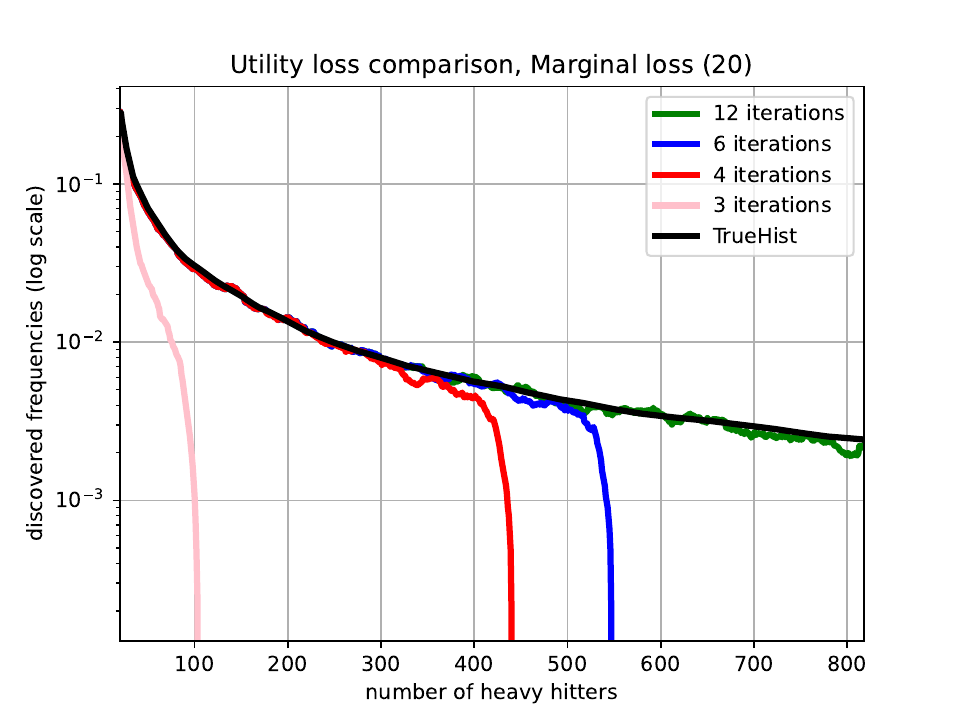}
    \caption{Marginal discovered frequencies}
    \label{fig:triehhmultitotal1}
     \end{subfigure}
     \hfill
     \begin{subfigure}{3in}
         \centering
         \includegraphics[width=3in]{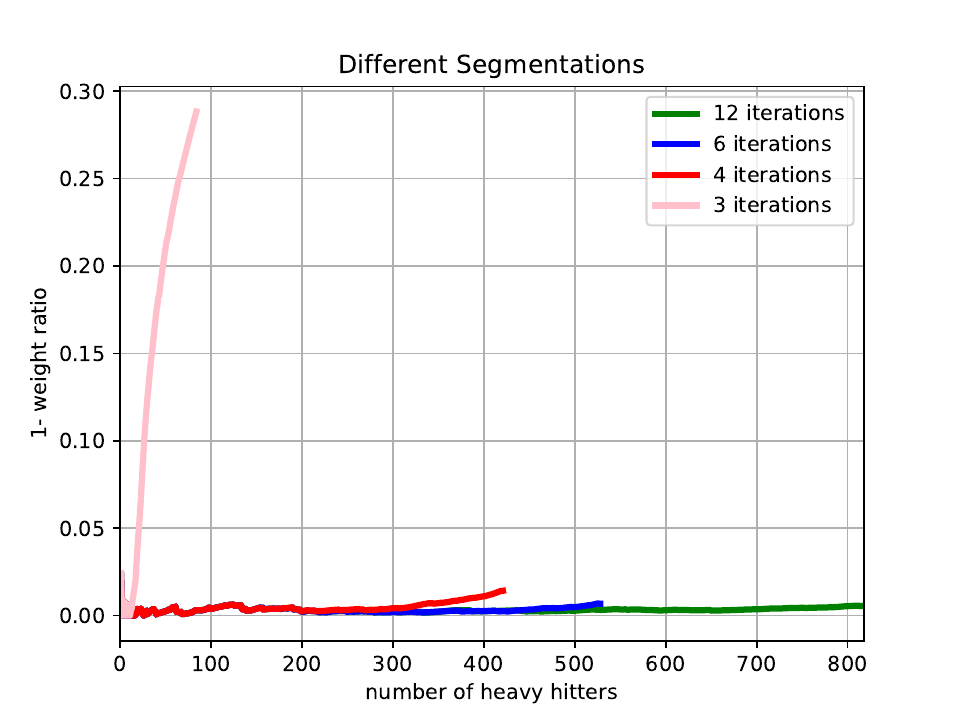}
    \caption{Total $\weightratio$ loss}
    \label{fig:triehhmultifreq1}
     \end{subfigure}
    \caption{The effect of different number of iterations on $\triehhg$ for multiple data points setting with weighted sampling($\epsagg=1$, $\delta=10^{-6}$)}
 \end{figure}

The total number of heavy hitters detected by $\triehhg$ algorithm, when the number of iterations are 12 (1 char), 6 (2 char), 4 (3 char), 3 (4 char), are $[816, 706, 506, 110]$ respectively. In this setting, 12 iterations shows the best utility. In Figure~\ref{fig:triehhmultitotal1} and ~\ref{fig:triehhmultifreq1} we used weighted sampling described in the original paper.

\begin{figure}[htb]
     \centering
          \begin{subfigure}{3in}
         \centering
         \includegraphics[width=3in]{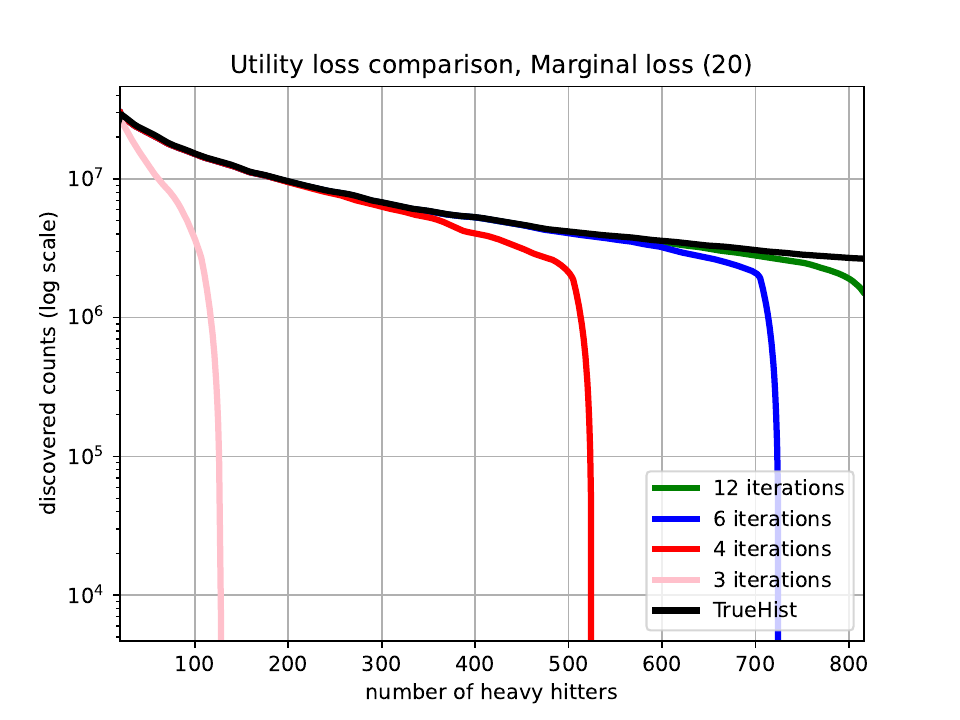}
    \caption{Marginal discovered counts}
    \label{fig:triehhmulticounts1}
     \end{subfigure}
     \hfill
     \begin{subfigure}{3in}
         \centering
         \includegraphics[width=3in]{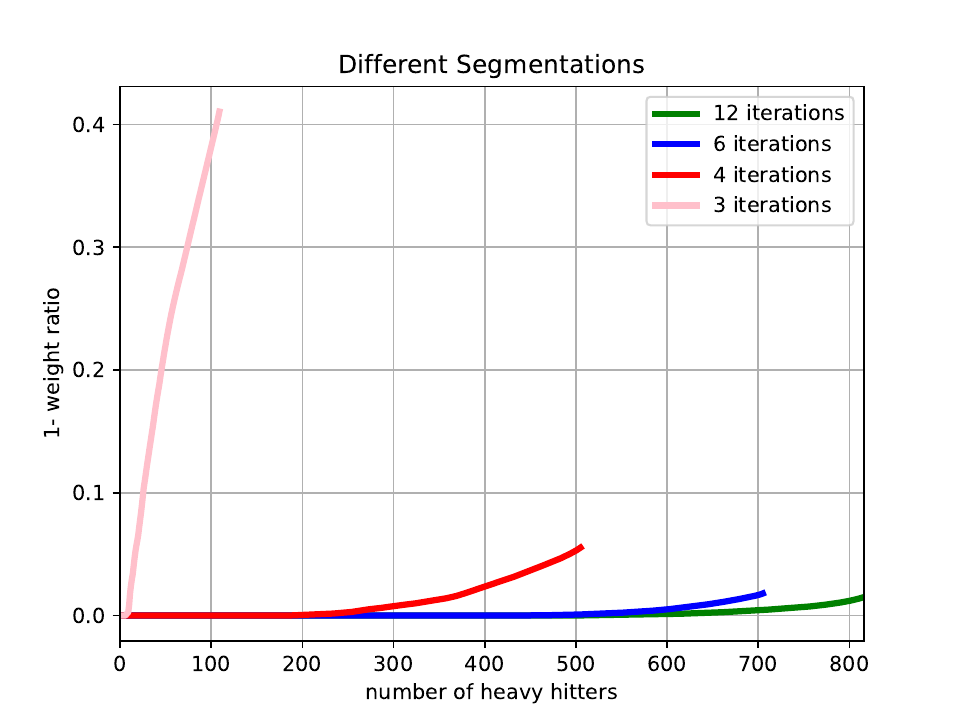}
    \caption{Total $\weightratio$ loss}
    \label{fig:triehhmultitotalunweighted}
     \end{subfigure}
     \caption{The effect of different number of iterations on $\triehhg$ for multiple data points setting with unweighted sampling ($\epsagg=1$, $\delta=10^{-6}$)}
 \end{figure}

To further improve the utility of $\triehhg$, we used unweighted sampling in another set of experiments. This new sampling scheme leads to finding $[816, 529, 422, 85]$ heavy hitters when having 12 (1 char), 6 (2 char), 4 (3 char), 3 (4 char) iterations respectively. Figure ~\ref{fig:triehhmulticounts1} and ~\ref{fig:triehhmultitotalunweighted} shows the loss and marginal discovered counts based on the unweighted sampling scheme. As demonstrated in these figures, unweighted scheme is able to find more heavy hitters and cause less utility degradation. Also using both sampling schemes, 12 iterations shows the highest utility. Thus, for the comparisons with $\ouralgorithm$ we use unweighted sampling, prefix-list and 12 iterations for $\triehh$.  

\section{TrieHH++}
\label{appendix:TrieHH++}
Based on Lemma 3 of~\cite{cormode2022sample}, $\triehhp$ achieves $(\epsilon, \delta)$ differential privacy when sampling rate $p_s = \alpha (1-e^{-\epsilon})$ where $ 0 < \alpha \le 1$ and $\epsilon < 1$ for $\delta = e^{-C_{\alpha}\theta}$, where $C_{\alpha}= \ln{1/\alpha}-1/(1+\alpha)$. The $\epsilon$ here is for one iteration. We used advanced composition in theorem 3.4 of~\cite{Kairouz:2017} to find the optimal $\epsilon$ per iteration which gives us $\epsagg$ of 1. We set the $\alpha$ parameter so that $\delta = 10^{-6}$. $\triehhp$ provide the analysis that shows the trade off between sampling and threshold values. We change the threshold $\theta$ from 10 to 20 and its effect on sampling rate are reported in Table~\ref{tab:TrieHH++} and~\ref{tab:TrieHH++large}.
\begin{table}[]
    \centering
    \resizebox{\columnwidth}{!}{%
    \begin{tabular}{|c|c|c|c|c|c|c|c|c|c|c|c|c|c|}
    \hline
        $\epsagg$ & \multicolumn{4}{|c|}{$1$} & \multicolumn{4}{|c|}{$0.5$} & \multicolumn{4}{|c|}{$0.25$} \\ \hline
        $T$ & $12$ & $6$  & $4$ & $3$  & $12$ & $6$  & $4$ & $3$& $12$ & $6$  & $4$ & $3$\\ \hline
         Sampling Rate  & $0.0071$ & $0.0129$ & $0.0193$& $0.0255$ & $0.0032$ & $0.0067$ & $0.0102$& $0.0138$   &$0.0016$ & $0.0034$ & $0.0053$& $0.0071$  \\ \hline
    \end{tabular}
    }
    \caption{Number of rounds effect on the sampling rate of $\triehhp$ ($\delta=10^{-6}$, $\numusers = 1.6\times10^6, \theta = 10$)}
    \label{tab:TrieHH++}
\end{table}

\begin{table}[]
    \centering
    \resizebox{\columnwidth}{!}{%
    \begin{tabular}{|c|c|c|c|c|c|c|c|c|c|c|c|c|c|}
    \hline
        $\epsagg$ & \multicolumn{4}{|c|}{$1$} & \multicolumn{4}{|c|}{$0.5$} & \multicolumn{4}{|c|}{$0.25$} \\ \hline
        $T$ & $12$ & $6$  & $4$ & $3$  & $12$ & $6$  & $4$ & $3$& $12$ & $6$  & $4$ & $3$\\ \hline
         Sampling Rate  & $0.0153
$ & $0.0305$ & $0.0449$  & $0.0589$ & $0.0078$ & $0.0159$ & $0.0239$&  $0.0319$   
         &$0.0039$ & $0.0082$ & $0.0123$& $0.0166$  \\ \hline
    \end{tabular}
    }
    \caption{Number of rounds effect on the sampling rate of $\triehhp$ ($\delta=10^{-6}$, $\numusers = 1.6\times10^6, \theta = 20$)}
    \label{tab:TrieHH++large}
\end{table}

\subsection{Single Data Point Setting for $\triehhp$}

\begin{figure*}
     \centering
     \begin{subfigure}{2in}
         \centering
         \includegraphics[width=2in]{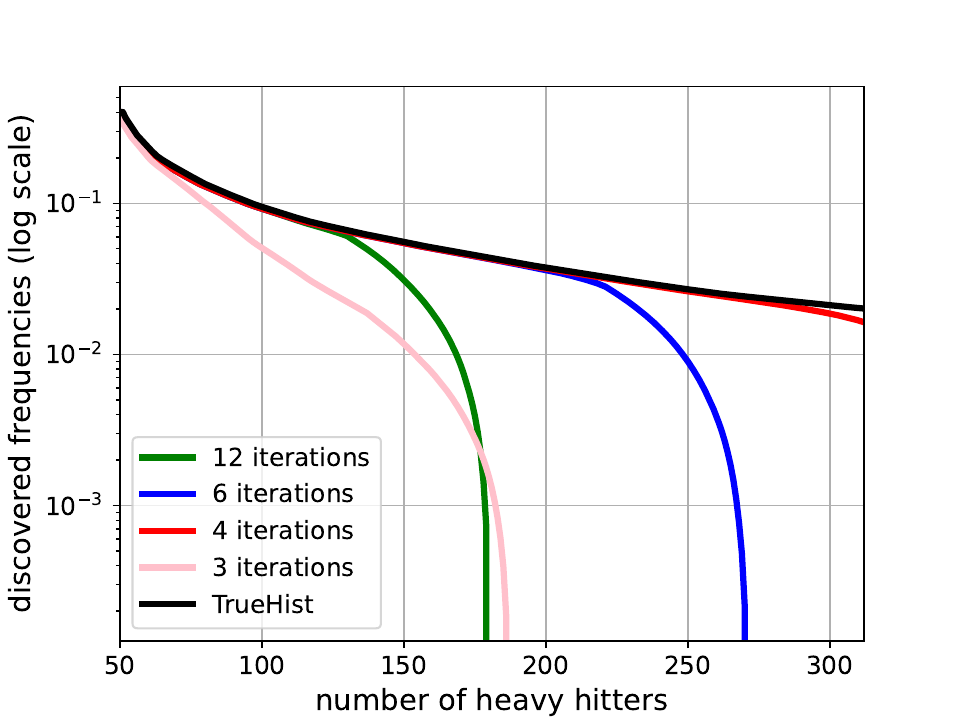}
    \caption{Marginal discovered frequencies}
    \label{fig:triehh+frequency}
     \end{subfigure}
     \hfill
          \begin{subfigure}{2in}
         \centering
         \includegraphics[width=2in]{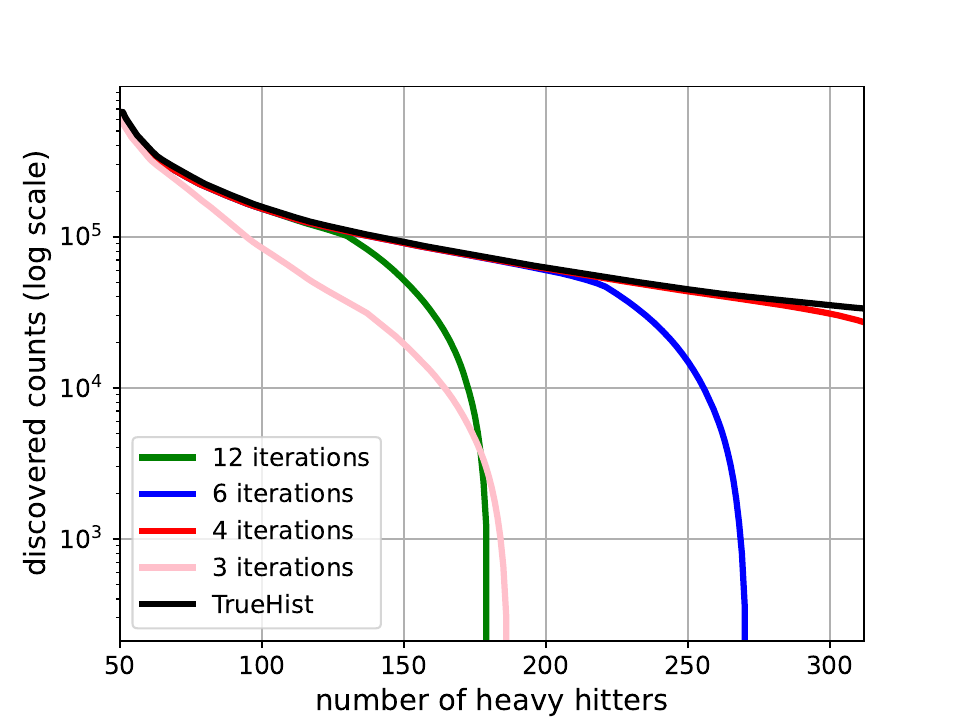}
    \caption{Marginal discovered counts}
    \label{fig:triehh+counts}
     \end{subfigure}
     \hfill
     \begin{subfigure}{2in}
         \centering
         \includegraphics[width=2in]{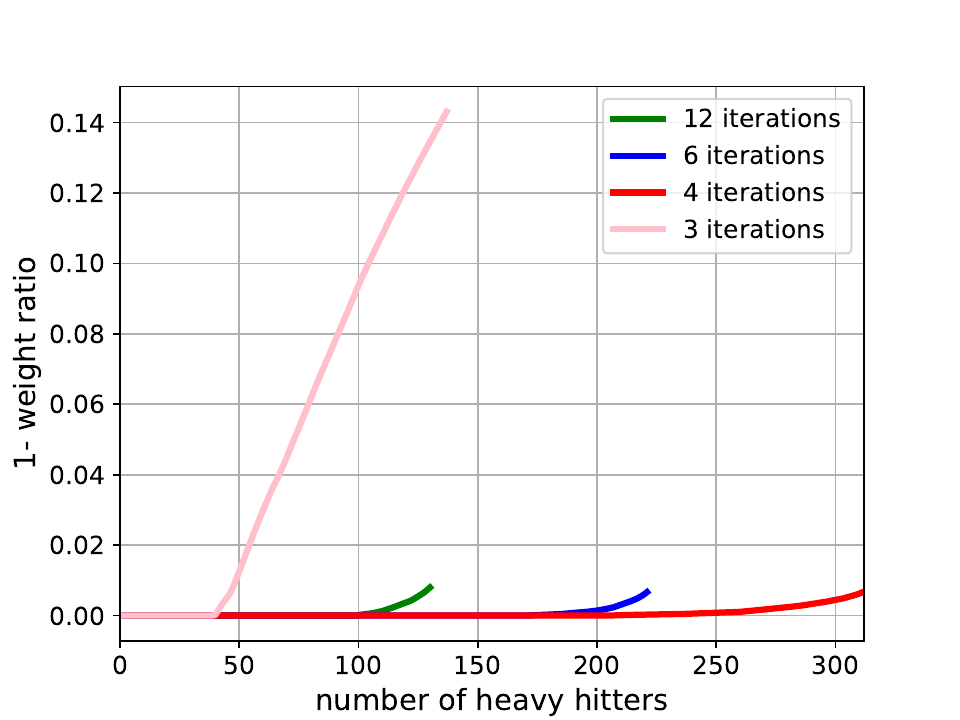}
    \caption{Total $\weightratio$ loss}
    \label{fig:triehh+total}
     \end{subfigure}
    \caption{The effect of different number of iterations on $\triehhp$ for single data point setting ($\epsagg=1$, $\delta=10^{-6}$)}
    \label{fig:triehh+single}
 \end{figure*}

In this section we discuss the effect of number of iterations on the utility of a single data point setting for $\triehhp$~\cite{cormode2022sample}. In Figure~\ref{fig:triehh+single}, we show the effect of different segmentation on the utility of the algorithm for a single data point per \device setting. For these experiments we used $\epsagg = 1$ and sampling rates are set based on table~\ref{tab:TrieHH++}. To evaluate the effect of different segmentation in this part we used the $\complimit = 10^7$ on the dimension. The number of heavy hitters detected by $\triehhp$ algorithm, when the number of iterations are 12 (1 char), 6 (2 char), 4 (3 char), 3 (4 char) are $[124, 223, 314, 137]$. Similar to $\triehhg$ having larger segments help with the algorithm since less number of iterations are required and consequently sampling rate becomes larger. However, by increasing the segment size to a certain point, the utility drops. The reason is by enlarging the segment length the number of prefixes that can be kept in each iteration reduces because of the dimension constraint. Thus, for the comparisons with $\ouralgorithm$ we used the best configurations which is having 4 iterations. 

\subsection{Multiple Data Points Setting for $\triehhp$}

\begin{figure*}
     \centering
     \begin{subfigure}{3in}
         \centering
         \includegraphics[width=3in]{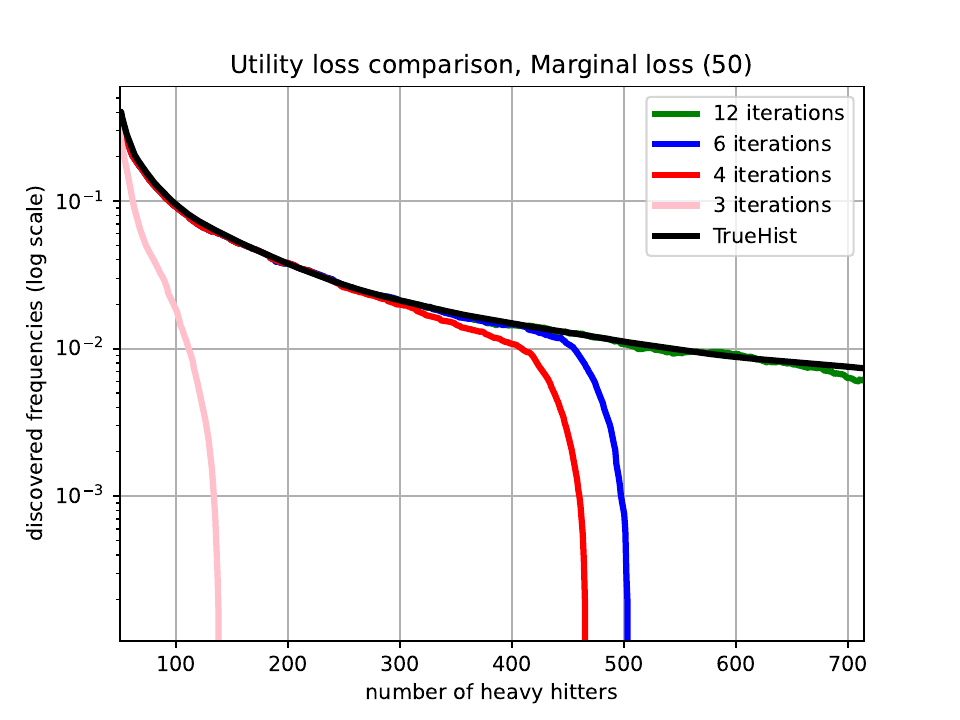}
    \caption{Marginal discovered frequencies}
    \label{fig:triehh+multitotal1}
     \end{subfigure}
     \hfill
     \begin{subfigure}{3in}
         \centering
         \includegraphics[width=3in]{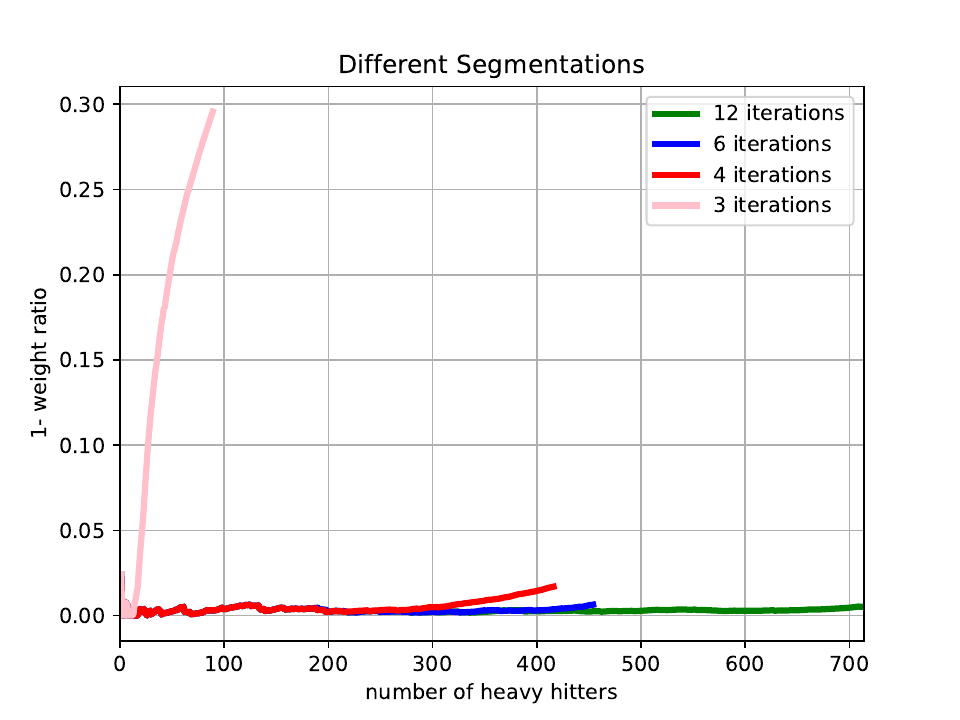}
    \caption{Total $\weightratio$ loss}
    \label{fig:triehh+multifreq1}
     \end{subfigure}
    \caption{The effect of different number of iterations on $\triehhp$ for multiple data points setting with weighted sampling($\epsagg=1$, $\delta=10^{-6}$)}
    \label{fig:triehh+multi}
 \end{figure*}

For our evaluation, we used the same binary encoding we described before. The total number of heavy hitters detected by $\triehhp$ algorithm, when the number of iterations are 12 (1 char), 6 (2 char), 4 (3 char), 3 (4 char), are $[714, 455, 417, 90]$ respectively. As shown in Figure~\ref{fig:triehh+multi}, in this setting, 12 iterations shows the best utility. 

\begin{figure*}
     \centering
          \begin{subfigure}{3in}
         \centering
         \includegraphics[width=3in]{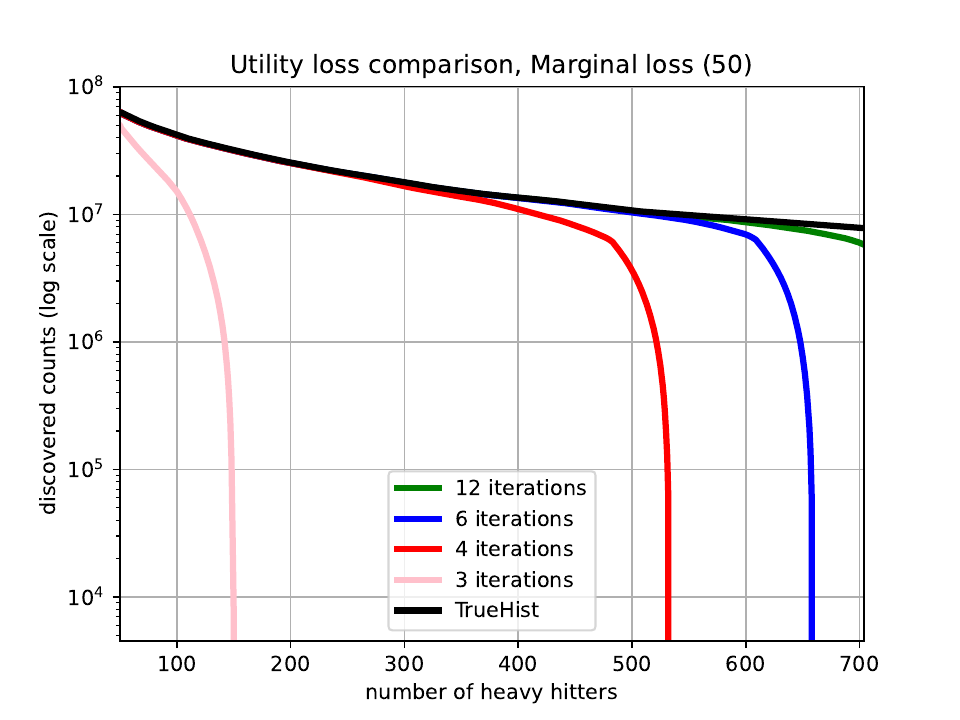}
    \caption{Marginal discovered counts}
    \label{fig:triehh+multicounts1}
     \end{subfigure}
     \hfill
     \begin{subfigure}{3in}
         \centering
         \includegraphics[width=3in]{Figure/TrieHH+CountsunWeighted.pdf}
    \caption{Total $\weightratio$ loss}
    \label{fig:triehh+multitotalunweighted}
     \end{subfigure}
     \caption{The effect of different number of iterations on $\triehhp$ for multiple data points setting with unweighted sampling ($\epsagg=1$, $\delta=10^{-6}$)}
 \end{figure*}

To further improve the utility of $\triehhp$, we used unweighted sampling in another set of experiments. This sampling scheme leads to finding $[704, 610, 484, 102]$ heavy hitters when having 12 (1 char), 6 (2 char), 4 (3 char), 3 (4 char) iterations respectively. Figure ~\ref{fig:triehh+multicounts1} and ~\ref{fig:triehh+multitotalunweighted} shows the loss and marginal discovered counts based on the unweighted sampling scheme. As demonstrated in these figures, unweighted scheme is able to find more heavy hitters and cause less utility degradation. Also using both sampling schemes, 12 iterations shows the highest utility. Thus, for the comparisons with $\ouralgorithm$ we use unweighted sampling, and 12 iterations for $\triehhp$ and we refer to it as $\opttriehhp$.

\section{Comparison of $\triehh$, $\opttriehhp$, and $\ouralgorithm$}\label{appendix:comparison}
\begin{figure*}
     \centering
     \begin{subfigure}{2in}
         \centering
         \includegraphics[width=2in]{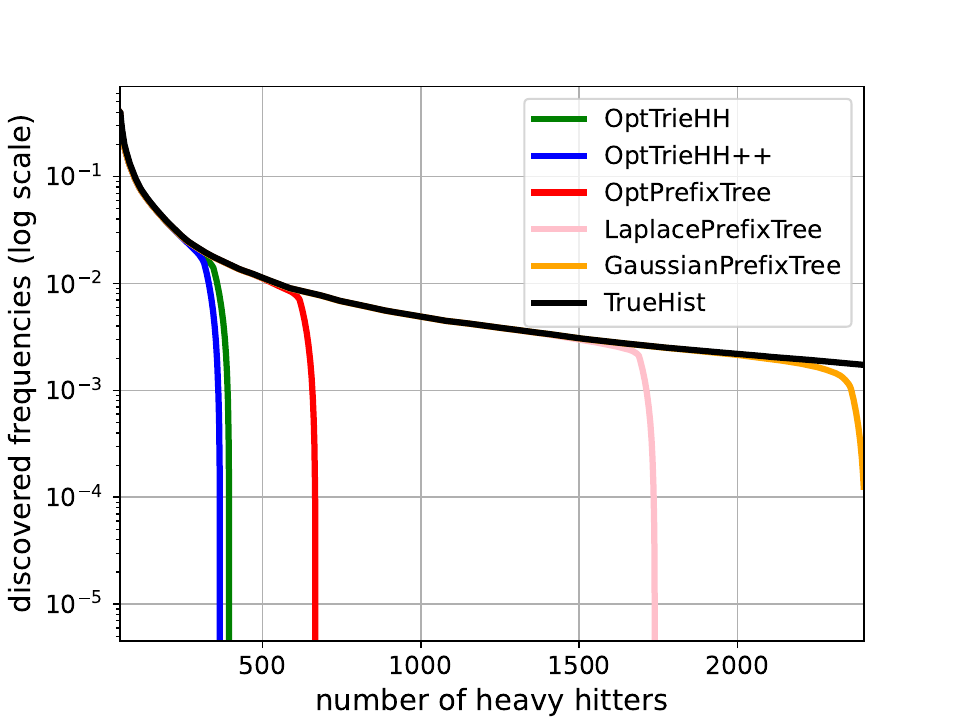}
\caption{Marginal discovered frequencies}
    \label{fig:compfreq}
     \end{subfigure}
     \hfill
     \begin{subfigure}{2in}
         \centering
         \includegraphics[width=2in]{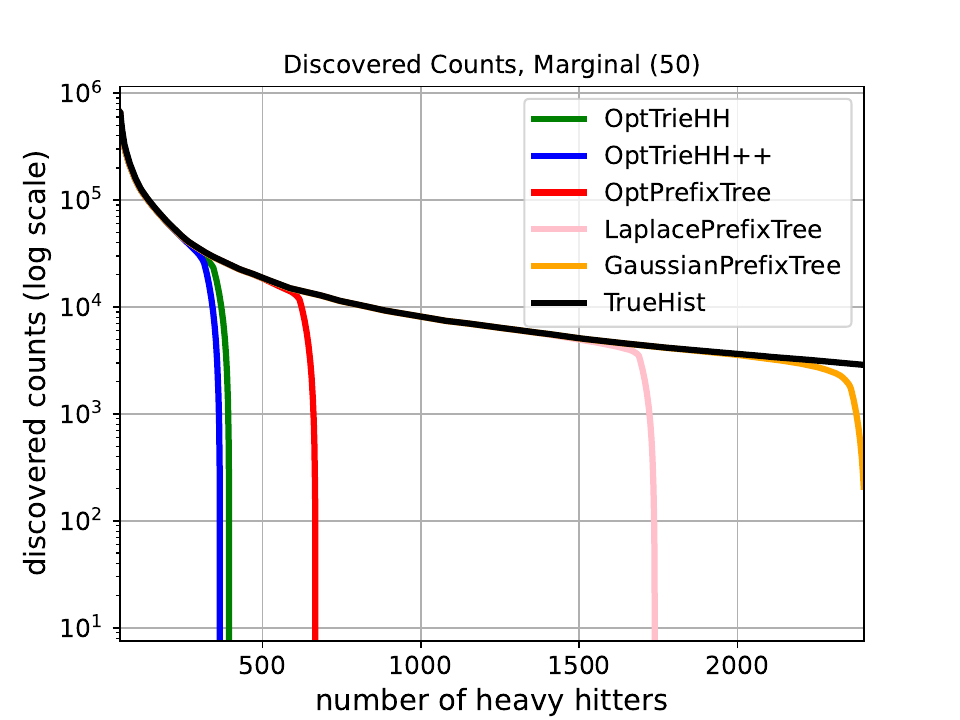}
    \caption{Marginal discovered counts }
    \label{fig:compcounts}
     \end{subfigure}
     \hfill
     \begin{subfigure}{2in}
         \centering
         \includegraphics[width=2in]{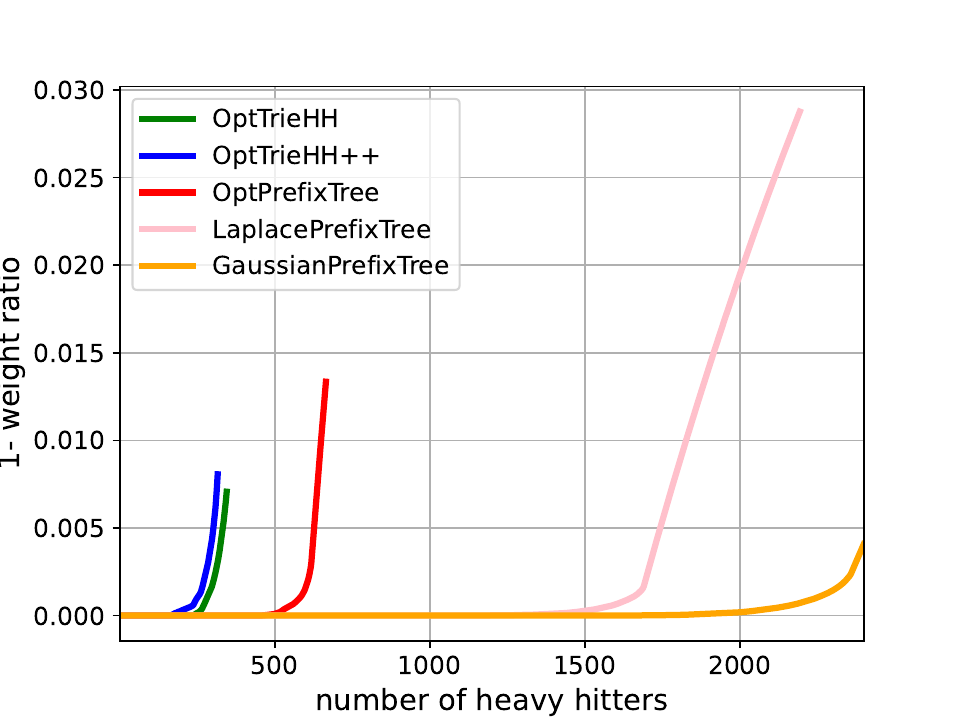}
\caption{Total $\weightratio$ loss }
    \label{fig:comptotal}
     \end{subfigure}
    \caption{Utility comparison between $\triehh$ and $\ouralgorithm$ for single data point setting($\epsagg=1$, $\delta=10^{-6}$, $\numiters=4$)}
 \end{figure*}

Zhu et al.'s primary experiments are on learning $n$-grams (words or length $n$ sentences) where they propose setting the segmentation length to be a single character. However, our analysis indicates that, in the single data point setting, larger segments provide the higher utility. Earlier in section~\ref{appendix:TrieHH} we showed the effect of different number of iterations on the sampling rate and utility of $\triehhg$. In this set of experiments we used 4 iterations that shows the highest utility for $\triehhg$ based on the dimension limitations ($\complimit = 10^7$). We refer to this optimized version of $\triehhg$ as $\triehh$. The same observation holds for $\triehhp$ and hence we use the same configuration for this algorithm. We refer to the optimized version of $\triehhp$ as $\opttriehhp$.

To have a fair comparison, we use the same binary encoding for all the three models. This binary encoding uses $5$ bits to convert English letters to a binary representation. In these experiments, $\totlen = 60$. We set $\falseratio = 2$ for $\ouralgorithm$. Using this method, $\ouralgorithm$ needs 4 iteration of unknown dictionary and uses the segmentation of $[23, 13, 12, 12]$. Figure~\ref{fig:compfreq} shows the marginal frequencies of discovered bins on y-axis and number of heavy hitters in x-axis. This figure shows the marginal value with sliding window of $50$ on y-axis. In conclusion, with the same dimension limit and binary encoding, our method is able to outperform $\triehh$ and $\opttriehhp$ by finding $1.85X$ and $2.1X$ more heavy hitters. Figure~\ref{fig:compcounts} shows the same plot but in the y-axis we have the marginal counts of discovered bins. Also, Figure~\ref{fig:comptotal} shows the total utility loss.
\section{Additional Experimental Results}\label{appen:expts-details}
\paragraph{Effect of Uniform vs Non-uniform Segmentation in Single Data Point Setting}

\begin{figure*}
     \centering
     \begin{subfigure}{3in}
         \centering
         \includegraphics[width=3in]{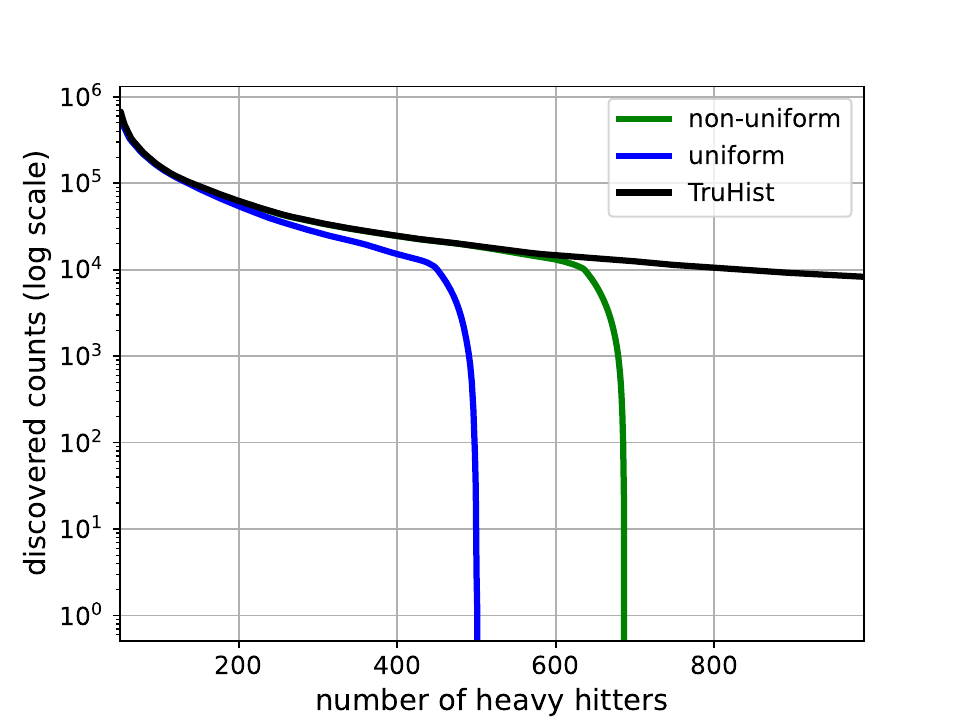}
    \caption{Marginal discovered counts comparison  }
    \label{fig:segmentationcounts}
     \end{subfigure}
     \hfill
     \begin{subfigure}{3in}
         \centering
         \includegraphics[width=3in]{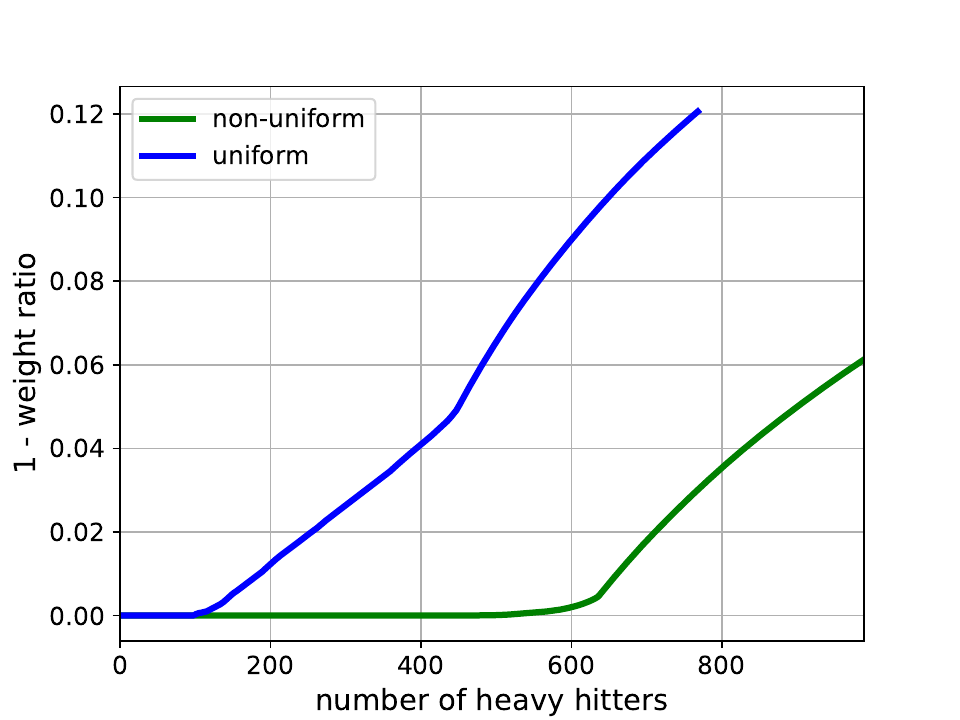}
    \caption{Total $\weightratio$ loss }
    \label{fig:segmentationtotal}
     \end{subfigure}
     
     \caption{Effect of different segmentation on $\ouralgorithm$ for single data point setting ($\epsagg=1$, $\delta=10^{-6}$, $\numiters=4$)}
     \label{fig:segmentation}
 \end{figure*}

As explained in~\ref{sec:expts} using our adaptive segmentation algorithm helps discovering more heavy hitters. In Figures~\ref{fig:segmentationcounts} and ~\ref{fig:segmentationtotal} we show the discovered counts and total utility loss comparison of this uniform and non-uniform segmentation. 
\paragraph{Effect of Dimension Limitation in Single Data Point Setting}

\begin{figure*}
     \centering
 
     \begin{subfigure}{3in}
         \centering
         \includegraphics[width=3in]{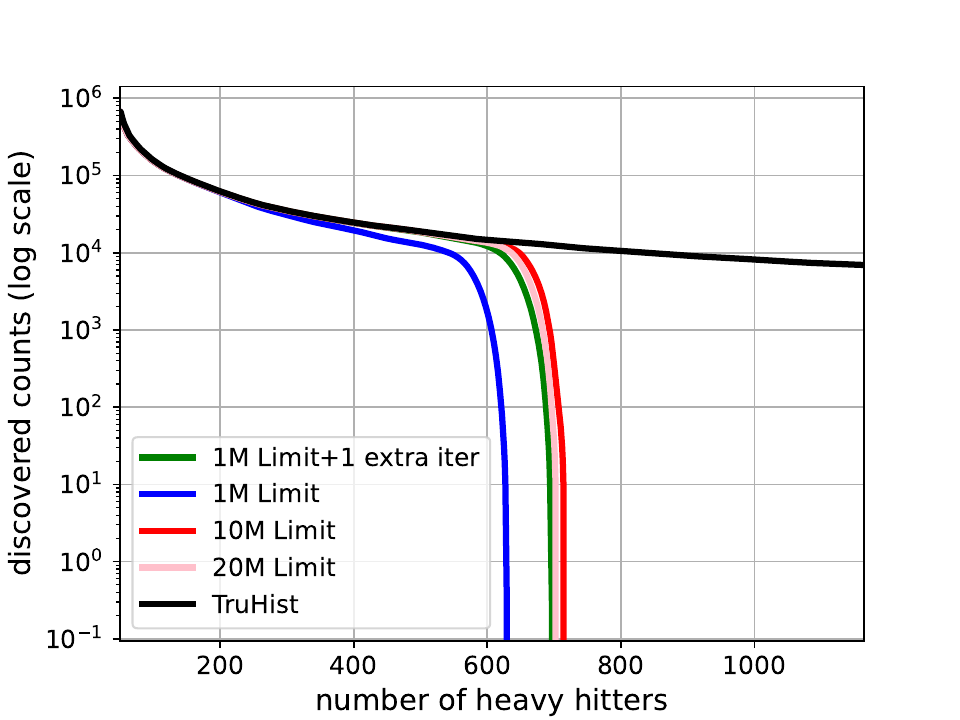}
     \caption{Marginal discovered counts}
    \label{fig:PayloadLimitCount}
     \end{subfigure}
     \hfill
    \begin{subfigure}{3in}
         \centering
         \includegraphics[width=3in]{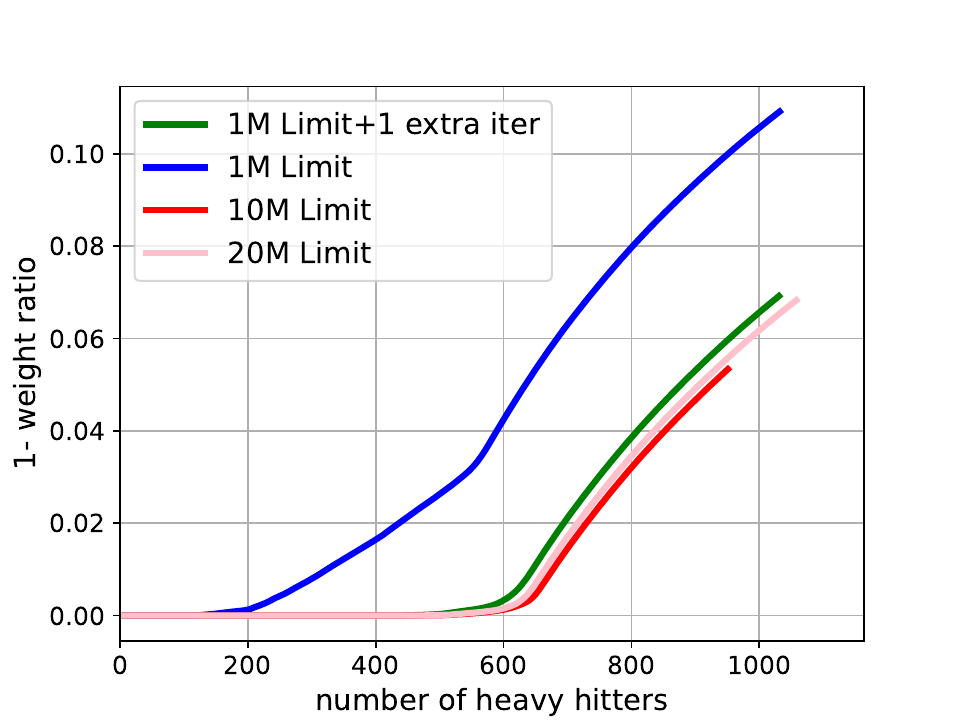}
     \caption{Total $\weightratio$ Loss}
    \label{fig:PayloadLimitTotal}
     \end{subfigure}
     \caption{Effect of dimension limitation for single data point setting on $\ouralgorithm$ utility ($\epsagg=1$, $\delta=10^{-6}$ and $T=4$)}   
 \end{figure*}

In Figures~\ref{fig:PayloadLimitCount} and ~\ref{fig:PayloadLimitTotal} we evaluated the effect of different dimension limitation parameters in the utility of the model. As shown, reducing the dimension limitation below a certain point, causes a significant utility drop. However, allowing the algorithm to have one extra iteration can help recovering top heavy hitters. 

\paragraph{Effect of Data Selection in Multiple Data Points Setting}

\begin{figure*}
     \centering
          \begin{subfigure}{2in}
         \centering
         \includegraphics[width=2in]{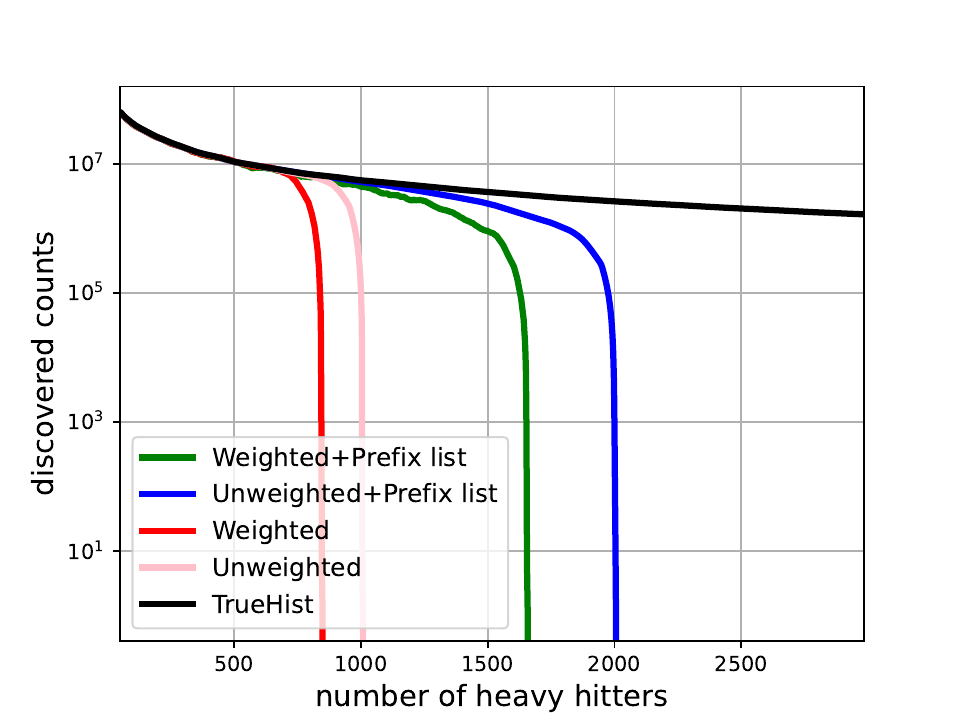}
    \caption{Marginal discovered counts }
    \label{fig:unweightcountsamp}
     \end{subfigure}
     \hfill
     \begin{subfigure}{2in}
         \centering
         \includegraphics[width=2in]{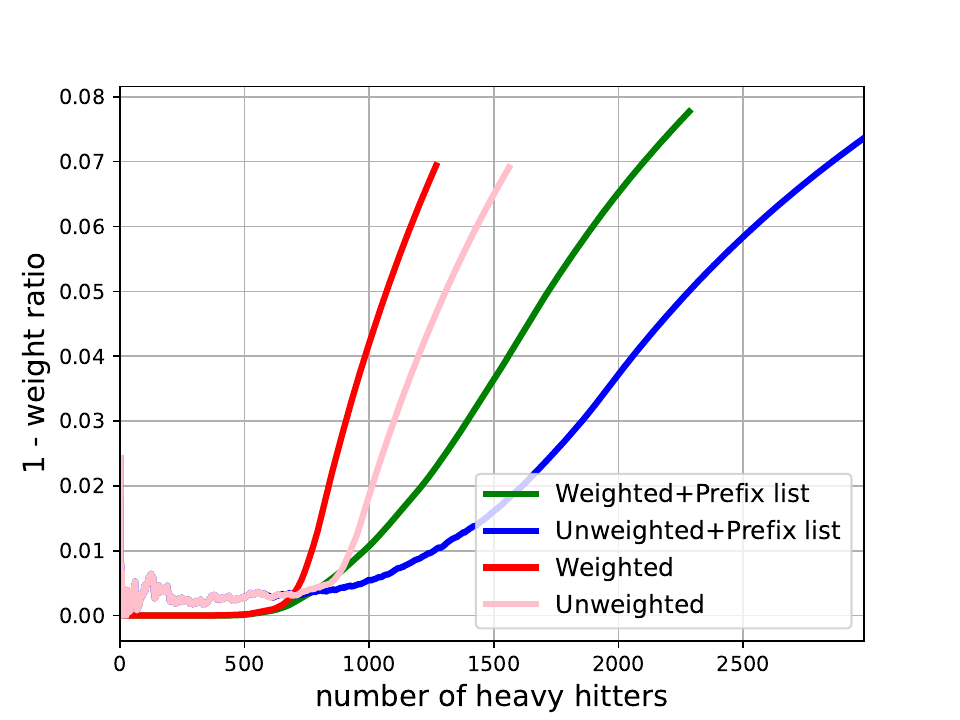}
    \caption{Total $\weightratio$ loss (weighted metric)}
    \label{fig:weighttotalsamp}
     \end{subfigure}
     \hfill
     \begin{subfigure}{2in}
         \centering
         \includegraphics[width=2in]{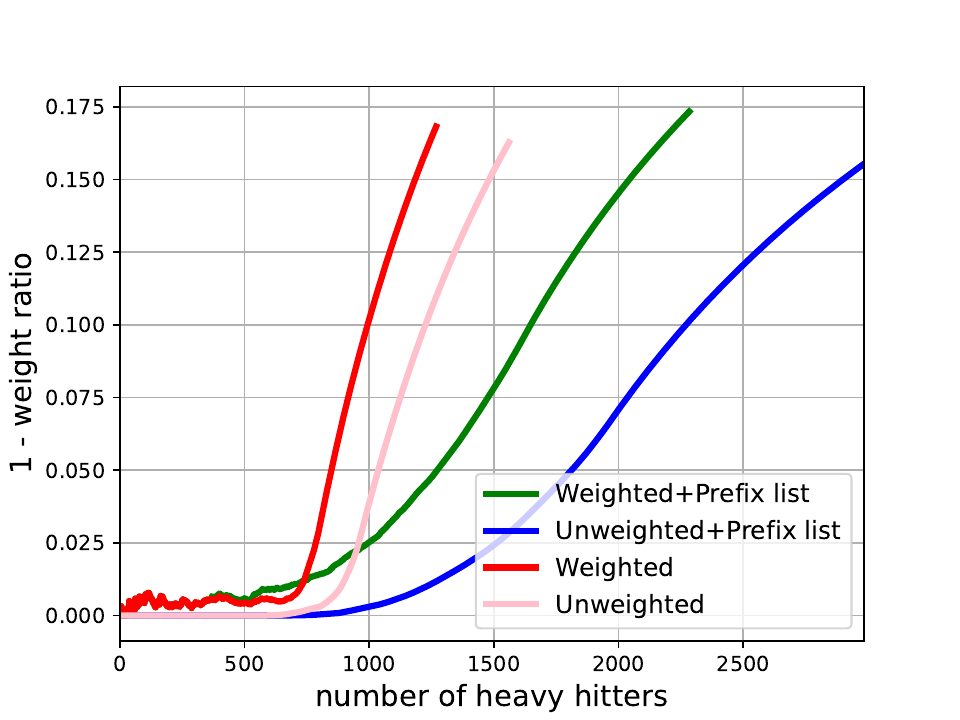}
    \caption{Total $\weightratio$ loss (unweighted metric) }
    \label{fig:unweighttotalsamp}
     \end{subfigure}
    \caption{Comparing different data selection schemes in $\ouralgorithm$ for multiple data points ($\epsagg=1$, $\delta=10^{-6}$, $\numiters=4$)}
    \label{fig:selection}
 \end{figure*}

There are different ways to measure the frequency of data points. In Section~\ref{sec:alg}, we discuss how averaging the distribution of words overall \devices can be used to define the global frequency of words (weighted metric). Figure~\ref{fig:weighttotalsamp} shows the total utility loss when using this distribution.

One other way to measure the frequency is the percentage of \devices who has the word in their support (unweighted metric). Figure~\ref{fig:unweightcountsamp} shows the global number of discovered bins based on how many \devices have the word. Figure~\ref{fig:unweighttotalsamp} demonstrates the total utility loss when using the average of users who has the data point as the frequency of words. As depicted using both distributions, unweighted data selection outperforms weighted data selection regardless of the frequency computation technique. Also prefix list benefits both of the data selection schemes. 

\paragraph{Effect of adding a \texttt{deny list} in Multiple Data Points Setting}

\begin{figure*}
     \centering
          \begin{subfigure}{2in}
         \centering
         \includegraphics[width=2in]{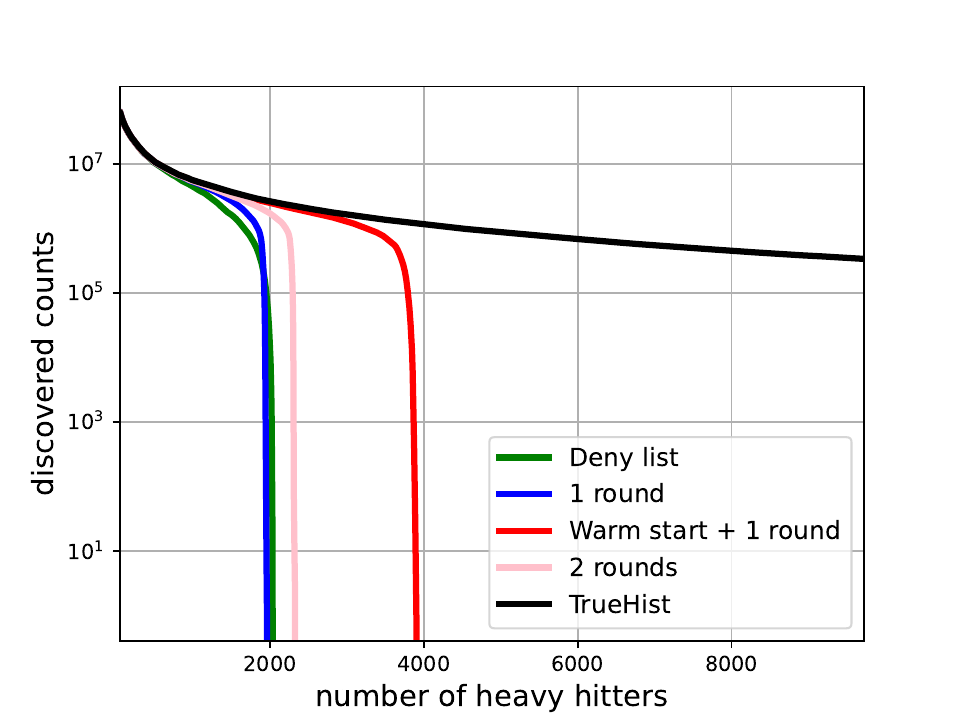}
    \caption{Marginal discovered counts }
    \label{fig:weightcountdeny}
     \end{subfigure}
     \hfill
          \begin{subfigure}{2in}
         \centering
         \includegraphics[width=2in]{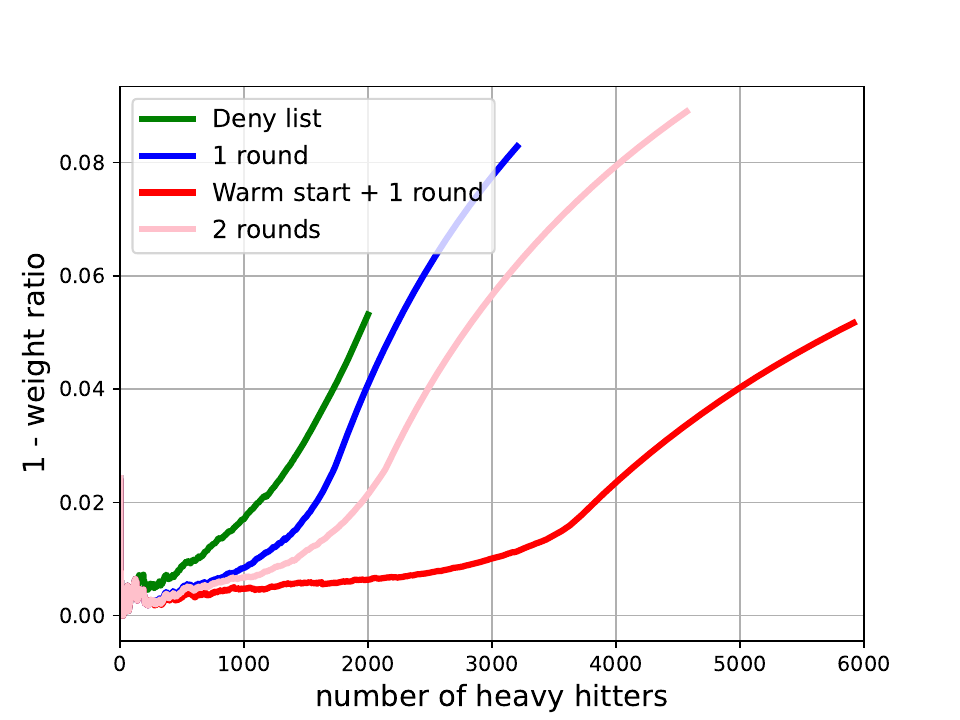}
    \caption{Total $\weightratio$ loss (weighted metric)}
    \label{fig:weightotaldeny}
     \end{subfigure}
     \hfill
     \begin{subfigure}{2in}
         \centering
         \includegraphics[width=2in]{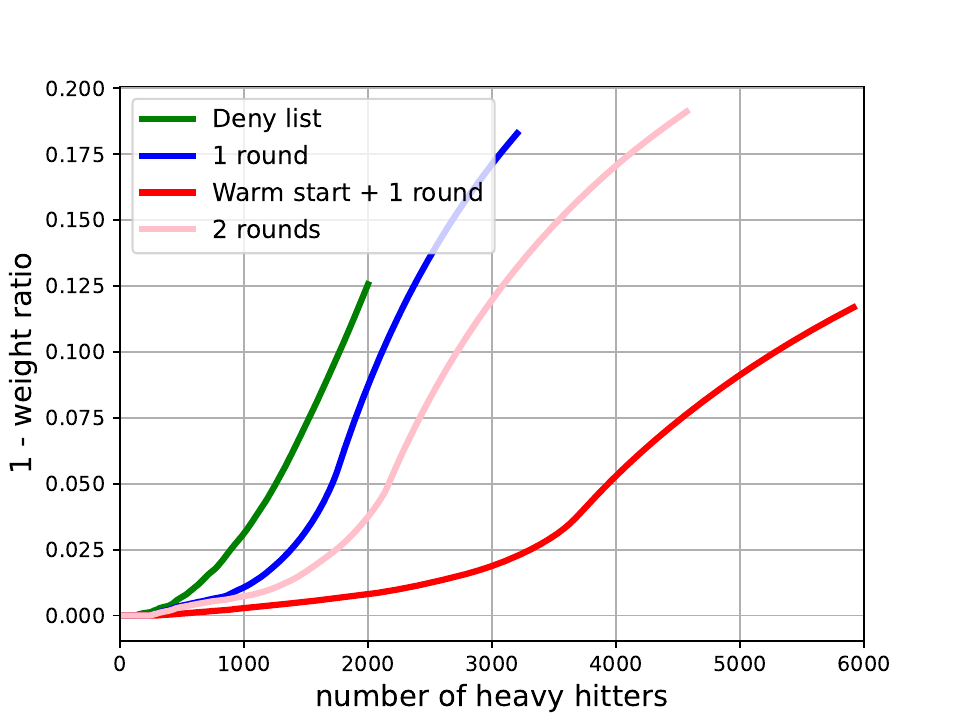}
    \caption{Total $\weightratio$ loss (unweighted metric)}
    \label{fig:unweightotaldeny}
     \end{subfigure}
     \caption{Effect of adding deny list to $\ouralgorithm$ for multiple data points setting ($\epsagg=1$, $\delta=10^{-6}$, $\numiters=4$)}
     \label{fig:deny}
 \end{figure*}

Figure~\ref{fig:weightotaldeny} shows the total utility loss when using the frequency of the words in \devices for extracting the global distribution. In Figure~\ref{fig:weightcountdeny} and ~\ref{fig:unweightotaldeny} we use the number of \devices with the word to demonstrate the true counts of the discovered words and total utility loss.  

\begin{figure*}
     \centering
     \begin{subfigure}{3in}
         \centering
         \includegraphics[width=3in]{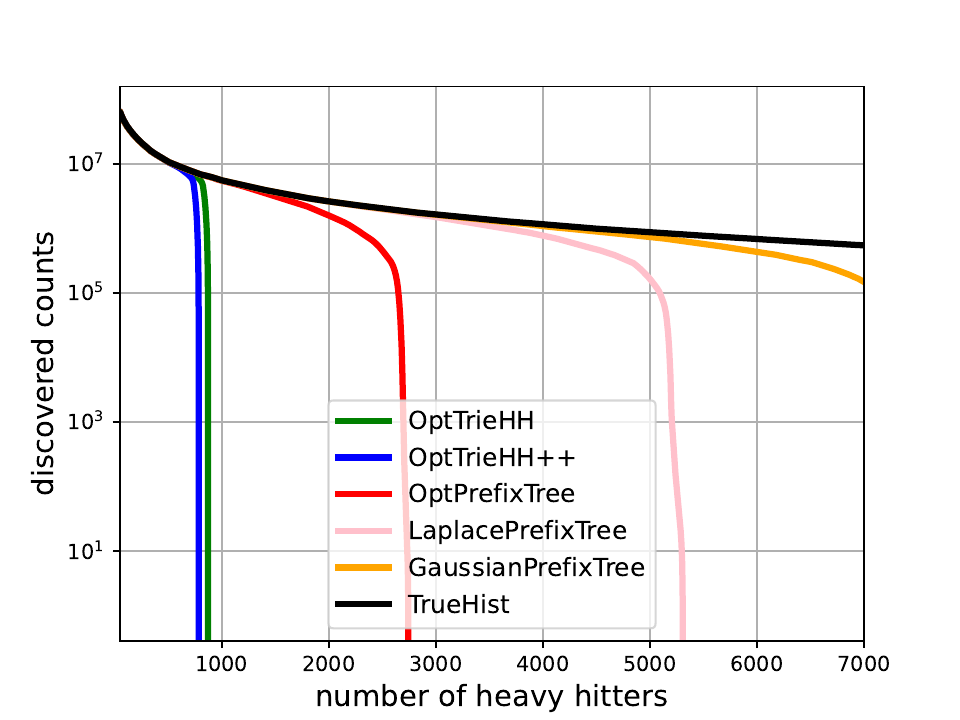}
    \caption{Marginal discovered counts }
    \label{fig:compcountmulti}
     \end{subfigure}
     \hfill
     \begin{subfigure}{3in}
         \centering
         \includegraphics[width=3in]{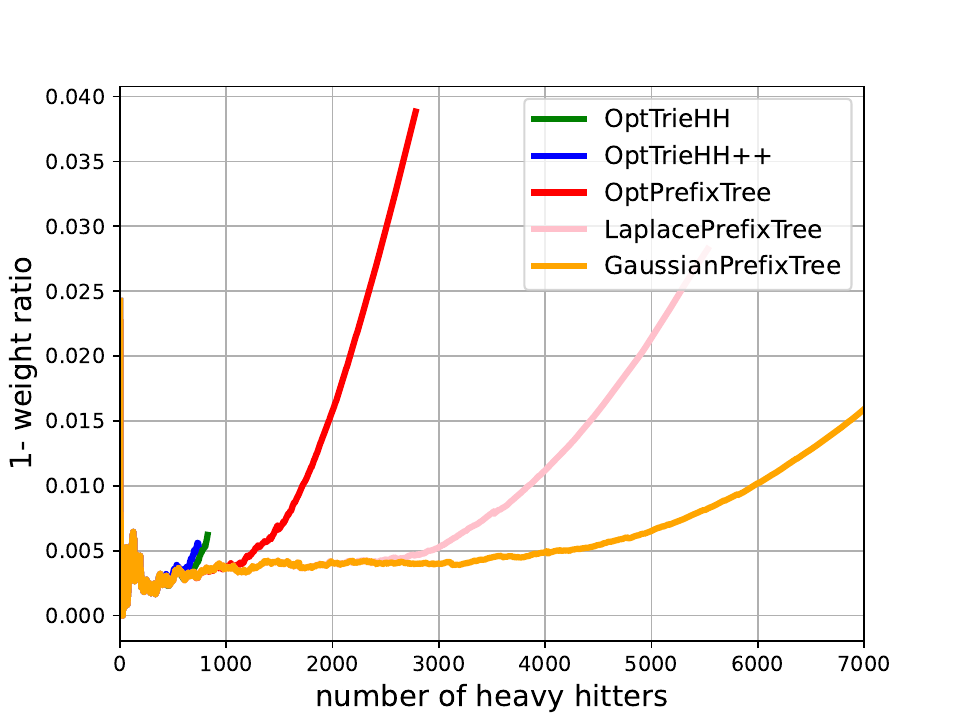}
    \caption{Total $\weightratio$ loss}
    \label{fig:compmulti}
     \end{subfigure}
     \caption{Comparing $\ouralgorithm$, $\triehh$, $\opttriehhp$ for multiple data points setting ($\epsagg=1$, $\delta=10^{-6}$, $\numiters=4$)}
 \end{figure*}

\paragraph{Comparison with previous works}
As illustrated in section~\ref{sec:expts} $\ouralgorithm$ outperforms $\triehh$ under the same constraints. In Figure~\ref{fig:compcountmulti} and~\ref{fig:compmulti}, we show a comparison of $\ouralgorithm$ and $\triehh$ for the multiple data points setting.

\end{document}